\documentclass[sigconf]{acmart}
\usepackage{tabularray}
\usepackage{tabularx}

\usepackage{enumitem}

\AtBeginDocument{%
  \providecommand\BibTeX{{%
    \normalfont B\kern-0.5em{\scshape i\kern-0.25em b}\kern-0.8em\TeX}}}

\setcopyright{rightsretained}
\copyrightyear{2024}
\acmYear{2024}
\acmDOI{10.1145/3664647.3681464}

\acmConference[MM'24]{Proceedings of the 32nd ACM International Conference
on Multimedia}{October 28 - November 1, 2024}{Melbourne, Australia.}
\acmISBN{979-8-4007-0686-8/24/10}

\begin{document}

\title[GPT4Video: A Unified Multimodal Large Language Model for Understanding and Generation]{\raisebox{-0.2ex}{\includegraphics[height=\ht\strutbox]{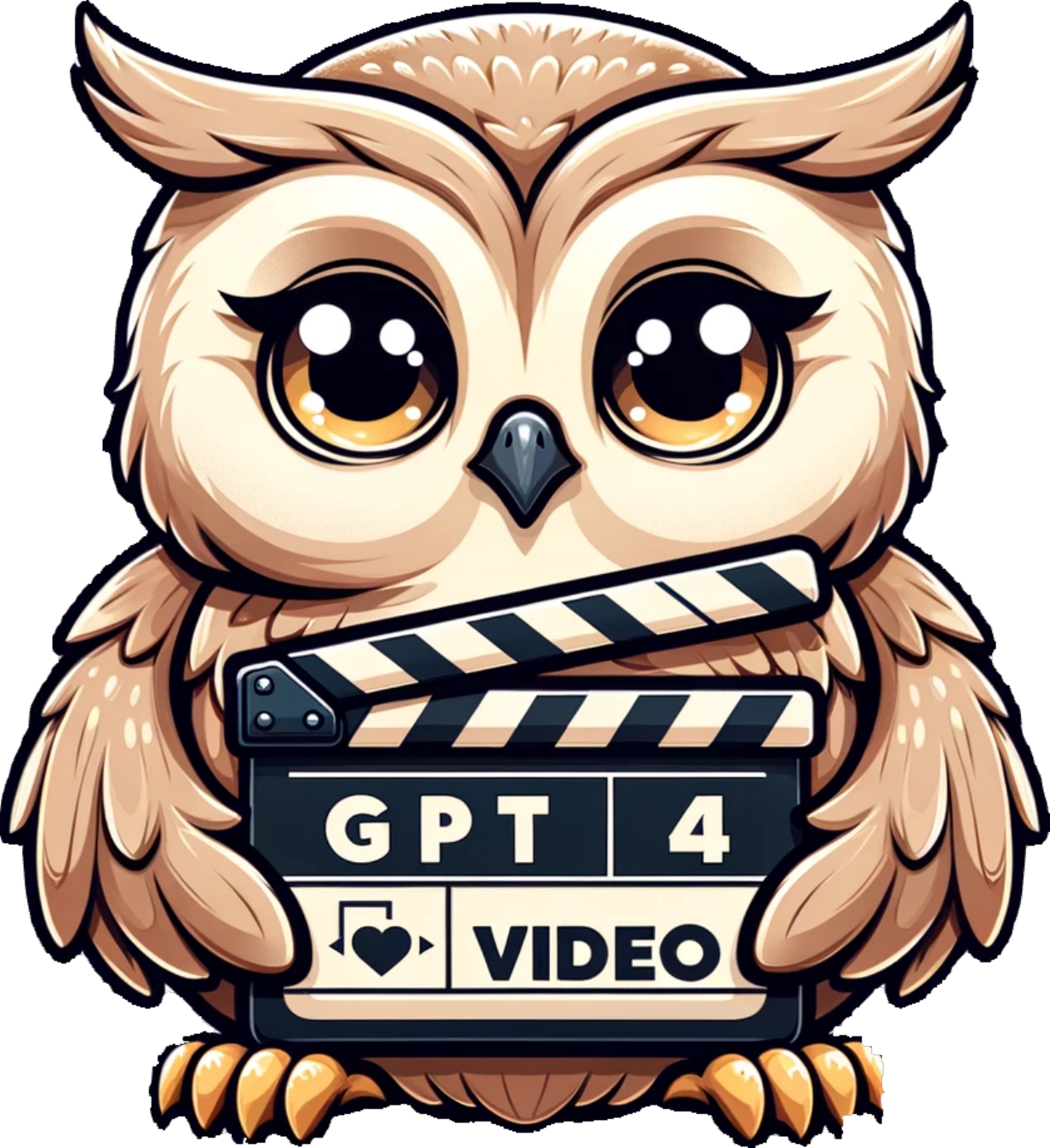}} GPT4Video: A Unified Multimodal Large Language Model for Instruction-Followed Understanding and Safety-Aware Generation}

% Zhanyu Wang, Longyue Wang, Zhen Zhao, Minghao Wu, Chenyang Lyu, Huayang Li, Deng Cai, Luping Zhou, Shuming Shi, Zhaopeng Tu
  % \streetaddress{1 Th{\o}rv{\"a}ld Circle}
  % \city{City}
  % \country{Country}
\author{Zhanyu Wang$^{1,2}$, Longyue Wang$^{1*}$, Zhen Zhao$^{1,2}$, Minghao Wu$^1$, Chenyang Lyu$^1$, Huayang Li$^1$, Deng Cai$^1$, Luping Zhou$^{2*}$, Shuming Shi$^1$, Zhaopeng Tu$^1$}
\affiliation{%
  \institution{$^1$Tencent AI Lab}
  \institution{$^2$The University of Sydney}
  \city{Shenzhen}
  \country{China}
  \authornote{Corresponding authors}
  }
\email{{zhanyu.wang, zhen.zhao, luping.zhou}@sydney.edu.au}
\email{{vinnylywang, mhaowu, georgelv, alanili, jcykcai, shumingshi, zptu}@tencent.com}
\email{https://gpt4video.github.io}


% \author{Zhanyu Wang}
% \affiliation{%
%   \institution{The University of Sydney}
%   % \streetaddress{1 Th{\o}rv{\"a}ld Circle}
%   \city{Sydney}
%   \country{Australia}}
% \email{xx@xx.xx}

% \author{Name}
% \affiliation{%
%   \institution{Institution}
%   % \streetaddress{1 Th{\o}rv{\"a}ld Circle}
%   \city{City}
%   \country{Country}}
% \email{xx@xx.xx}

% \author{Name}
% \affiliation{%
%   \institution{Institution}
%   % \streetaddress{1 Th{\o}rv{\"a}ld Circle}
%   \city{City}
%   \country{Country}}
% \email{xx@xx.xx}
\renewcommand{\shortauthors}{Zhanyu Wang et al.}

\begin{abstract}
Recent advances in Multimodal Large Language Models (MLLMs) have constituted a significant leap forward in the field, particularly in the processing of videos, which encompasses inherent challenges such as spatiotemporal relationships. However, existing MLLMs are predominantly focused on the comprehension of video inputs, with limited capabilities in generating video content. In this paper, we present \texttt{GPT4Video}, a unified framework that seamlessly and lightly integrates with LLMs, visual feature extractors, and stable diffusion generative models for cohesive video understanding and generation. Moreover, we explore a \texttt{text-only finetuning} approach to equip models for instruction-following and safeguarding in multimodal conversations, enhancing training efficiency and generalization capabilities. Additionally, we construct multi-turn and caption-interleaved datasets for finetuning and benchmarking MLLMs, which serve as solid resources for advancing this field. Through quantitative and qualitative assessments, \texttt{GPT4Video} demonstrates the following advantages: 1) The framework incorporates video generation ability without adding extra training parameters, ensuring seamless compatibility with various video generators. 2) The model achieves superior performances across a variety of benchmarks. For instance, it outperforms Valley~\cite{luo2023valley} by 11.8\% on video question answering, and surpasses NExt-GPT~\cite{nextgpt} by 2.3\% on text-to-video generation. 3) As safety pioneers in open-source MLLMs, we developed finetuning and evaluation datasets, securing an F1 score exceeding 80\% in blocking harmful content during understanding and generating videos. In general, \texttt{GPT4Video} shows potential to function as a real-life assistant, marked by its effectiveness, adaptability, and safety.
\end{abstract}





\maketitle

\section{Introduction}\label{sec:intro}
\begin{figure}[t]
    \centering
    \centerline{\includegraphics[width=0.9\linewidth]{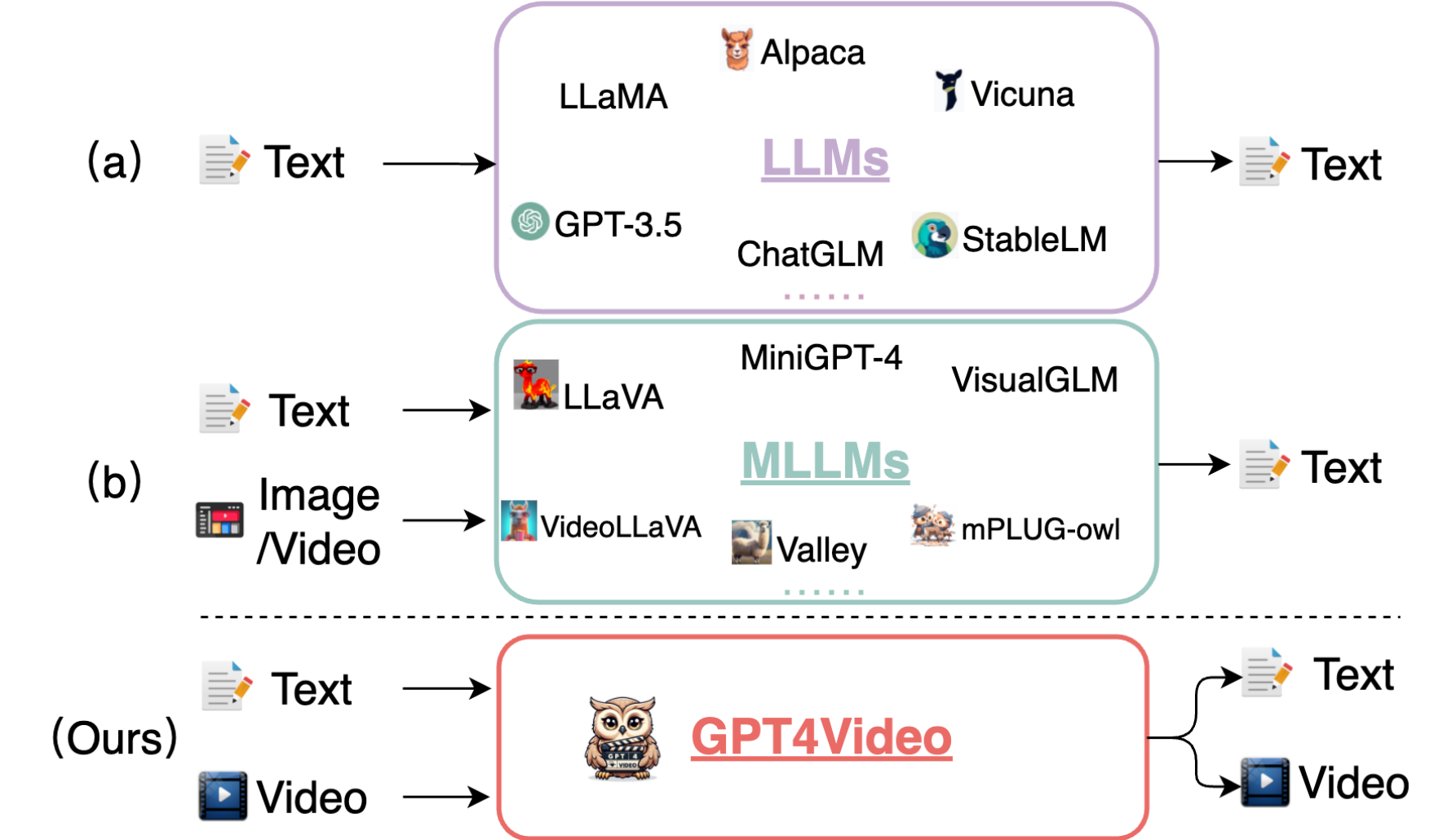}}
      \caption{Pipeline comparison of our GPT4Video with the current LLMs and MLLMs.}
    \label{fig:firstPage}
    \vspace{-5mm}
\end{figure}

{L}{arge} language models (LLMs) such as LLaMA~\cite{llama2}, ChatGLM~\cite{du2022glm}, Vicuna~\cite{vicuna2023} (in Figure~\ref{fig:firstPage} (a)), have undergone significant advancements, paving the way for general artificial intelligence (AGI) and showcasing remarkable zero-shot capabilities across diverse linguistic tasks. Based on this foundation, multimodal large language models (MLLMs) like MiniGPT-4~\cite{miniGPT-4}, Video-LLaMA~\cite{Video-llama}, Macaw-LLM~\cite{macaw-llm} (in Figure~\ref{fig:firstPage} (b)) have been successively introduced. These models are capable not only of processing textual information but also of handling data across various modalities such as images, audio, and videos, demonstrating exceptional performance in multimodal tasks~\cite{yin2023survey}. Nevertheless, current MLLMs primarily focus on processing {multimodal inputs}, yet they fall short in generating multimodal information on the output side. For a refined AGI system~\cite{goertzel2014artificial}, it is crucial not only to understand multimodal inputs but also to efficiently generate them, mirroring human interaction in real-world settings.

Nonetheless, endowing LLMs with multimodal generation capabilities presents a significant challenge: {\em what} content aligns with user requirements, and {\em when} to generate it appropriately.
Recent studies have made preliminary forays in this direction~\cite{gill, nextgpt, zheng2023minigpt5}. GILL~\cite{gill} initially investigated leveraging LLMs for image generation, introducing "generative vokens" (i.e. the special token ``\texttt{<IMG>}'') into the text vocabulary of LLMs.
NExt-GPT~\cite{nextgpt} further extends the concept of ``generative vokens" to a variety of modalities (e.g. image, audio, and video). Though the approaches have demonstrated certain efficacy, they have several inherent limitations:
\begin{enumerate}[leftmargin=*,topsep=0.1em,itemsep=0.1em,parsep=0.1em]
    \item The representational capacity of ``generative vokens" is limited. Due to the constraints of their training mechanism, they fail to fully harness the outstanding potential of LLMs in text generation and emergence ability. 
    \item Adding new vocabulary to the LLM's lexicon may disrupt the LLM's original capabilities. 
    \item These methods lack the flexibility to adapt different generation models. For instance, upgrading the text-to-image/video model necessitates retraining the LLM.
\end{enumerate}
Note that, quantitative analyses on the representative method~\cite{nextgpt} in Section~\ref{sec:experiments} provide support for the above claims.

To address the limitations, we present {\em GPT4Video}, a unified framework for augmenting LLMs with both video understanding and generation capabilities (in Figure~\ref{fig:firstPage} (Ours)). 
Inspired by previous success~\cite{mplug_owl}, we leverage the robust capabilities of pretrained LLM and visual feature extractor for {\em video understanding}. 
Considering the the inherent complexity of video over images, we introduce a video abstractor with a dual attention mechanism~\cite{zhao2017video_dual1,fu2019dual} to facilitate precise alignment between video and text. For {\em video generation}, instead of incorporating ``generative vokens'', we train LLMs to produce the video descriptions with tags at the appropriate moment. These tags and descriptions respectively act as trigger (``{\em when}'') and input prompts (``{\em what}'') for a pre-trained text-to-video model.

Previous studies commonly created and then finetuned MLLMs using multimodal instruction datasets for understanding abilities (e.g. text-video interleaved data) \cite{macaw-llm,chen2023sharegpt4v,chen2024allava}.
In contrast, we investigate a {\em text-only finetuning} method for video generation, replacing video with its caption in instruction datasets during the second-stage training.\footnote{The assumption is that the model has acquired vision-language knowledge at the first-stage training \cite{miniGPT-4,liu2023visual_llava}.} Accordingly, we construct datasets and apply the proposed finetuning for two scenarios: equipping models for instruction-following (45K instances) and safeguarding (4K instances).
Our {GPT4Video} has the following appealing advantages:
\begin{itemize}[leftmargin=*,topsep=0.1em,itemsep=0.1em,parsep=0.1em]
    \item {\bf Seamless Integration}: The framwork requires no additional modules or training parameters, thus can be seamlessly integrated into a variety of text-to-video models (e.g. ZeroScope~\cite{luo2023videofusion} and VideoCrafter~\cite{chen2023videocrafter1}). 
 
    \item {\bf Instruction-Followed Video Generation}: Textual data has more powerful abstraction and expression capabilities in summarizing content. The fine-tuning with our text-based instructions is expected to efficiently harness LLMs' generalization abilities in generating content-rich video prompts that align with instructions.
    
    \item {\bf Safeguarding}: Through a simple safety alignment, our model substantially rejects both the processing of harmful video inputs and the generation of harmful video outputs. This makes it one of the first models to incorporate safeguarding features among open-source MLLMs.\footnote{We primarily focus on explicit content, as pornography represents the most significant risk in the realm of video generation.} 
\end{itemize}

We conduct experiments on a variety of multimodal benchmarks, including open-ended question-answer~\cite{msr-vtt,msvd-qa}, video captioning~\cite{msr-vtt}, text-to-video generation~\cite{msr-vtt} and safeguarding (ours).
Results show that our model significantly and consistently outperforms the advanced models, NExt-GPT~\cite{nextgpt}, demonstrating the effectiveness and universality of our framework and approach. 
Our {\bf main contributions} are:
\begin{itemize}[leftmargin=*,topsep=0.1em,itemsep=0.1em,parsep=0.1em]
    \item We propose a unified framework that enhances LLMs with video understanding and generation capabilities through seamless and lightweight integration with pre-trained Abstractor and T2V models, demonstrating remarkable extensibility.
    
    \item We explore a simple and effective method, text-only finetuning, for developing instruction-followed models, showing substantial improvements in performance. As safety pioneers, we further employ this approach for safety alignment, offering an alternative to einforcement learning from human feedback (RLHF). 

    \item We build and release valuable datasets to facilitate future work on multimodal LLMs, including (1) a {instruction dataset} for video generation that covers a wide range of conversational scenarios; (2) a {safety benchmark} for evaluating and enhancing safeguarding capabilities in video-interleaved conversations.
\end{itemize}

%%%%-------------------------------------------Related Work--------------------------------------------------------------

\section{Related Work}
\label{sec:related_work}

\noindent\textbf{Multimodal Language Models}
Numerous studies have developed multimodal language models that can handle visual inputs and text outputs, or vice versa, such as \cite{FrozenBiLM,ceylan2023pix2video}. With these advancements of LLMs, some researches have focused on learning a joint embedding space for multiple modalities, as demonstrated in \cite{clip,DBLP:journals/corr/abs-2305-05665}. Others have combined pre-trained single-modality models to showcase impressive zero-shot capabilities \cite{DBLP:conf/nips/AlayracDLMBHLMM22, DBLP:journals/corr/abs-2301-12597}. More recently, there has been a growing interest in enabling multi-modal LLMs to follow instructions, as shown in \cite{miniGPT-4, mplug_owl, dai2023instructblip}. To facilitate research in this area, \cite{DBLP:journals/corr/abs-2212-10773} introduced MultiInstruct, the first multi-modal instruction tuning benchmark dataset covering a wide range of tasks and categories. Additionally, \cite{liu2023visual_llava} explored multi-modal instruction-tuning using machine-generated data, while \cite{macaw-llm} fine-tuned all model parameters to allow the textual LLM to process four modalities.

\noindent\textbf{Large Language Models}
Large language models (LLMs) commonly refer to as Transformer-based language models with billions of parameters \citep{vaswani2017attentionisallyouneed} and have revolutionized the research paradigm in natural language processing community \citep{du2022glm,llama2}. Furthermore, recent works have demonstrated that supervised fine-tuning, also known as instruction-tuning, can effectively improve the zero-shot performance of these LLMs \citep{alpaca,vicuna2023}. \citet{DBLP:journals/corr/abs-2303-18223} present a comprehensive survey on the research of LLMs.

\begin{figure*}[t]
    \centering
    \centerline{\includegraphics[width=0.75\linewidth]{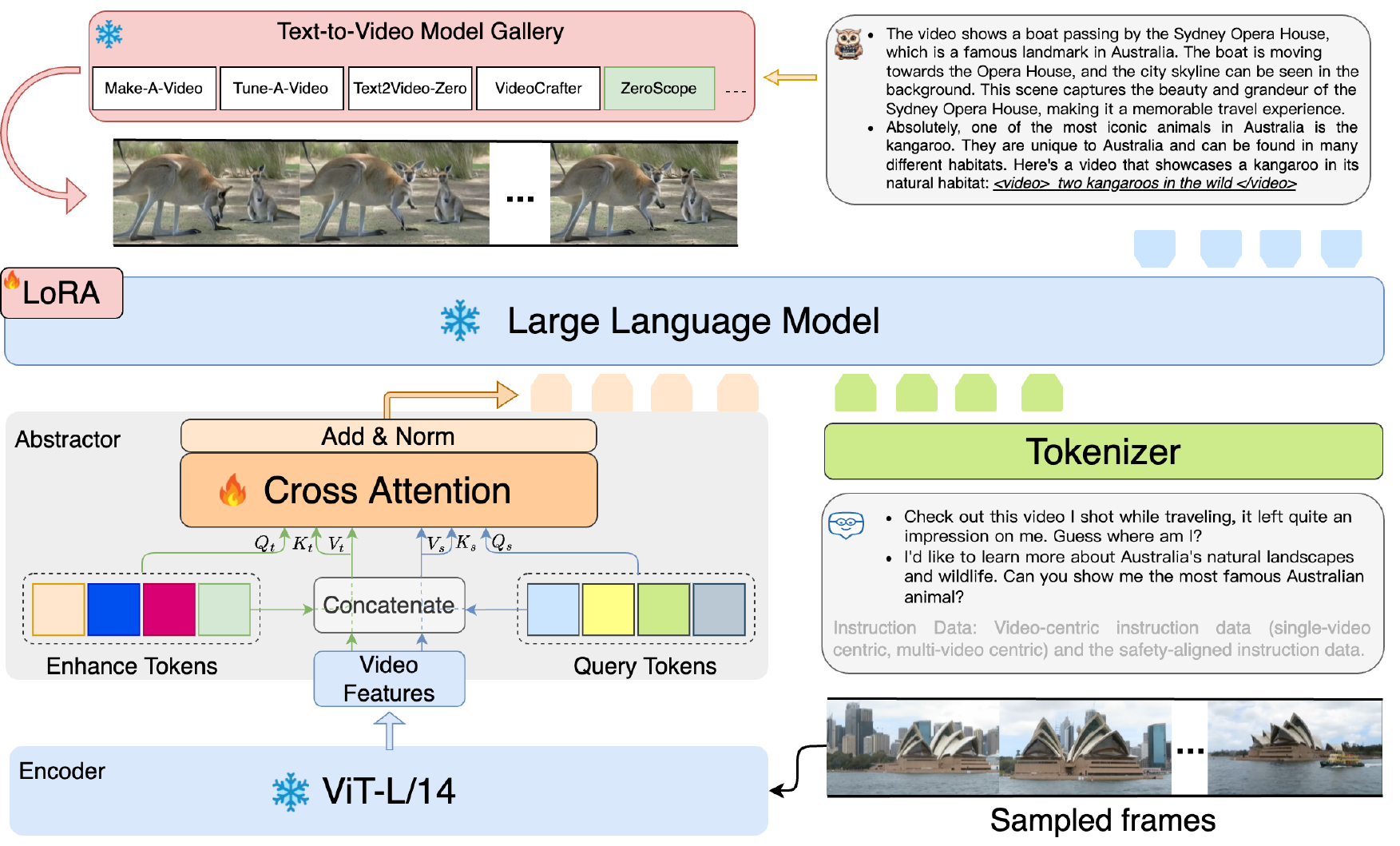}}
      \caption{Architectural Overview of GPT4Video. This framework has two components for video processing. The video encoding module employs a frozen ViT-L/14 model to capture raw video features, while the video abstraction module utilizes a transformer-based cross attention layer and two novel learnable tokens, designed to condense video features. The core of GPT4Video is powered by a frozen LLaMA model, efficiently fine-tuned via LoRA. The LLM is trained with custom video-centric and safety-aligned data, enabling it to comprehend videos and generate appropriate video prompts (indicated by underlined text). These prompts are then used to create videos from the Text-to-Video Model Gallery. Icons of a snowflake and a flame visually distinguish between the non-trainable and trainable parameters within the system. The two bullet points highlight GPT4Video's dual capabilities in understanding and generating video content.}
    \label{fig:framework}
    % \vspace{-1em}
\end{figure*}

\noindent\textbf{Text-to-Image/Video Generation}
Text-to-image/video generation refers to the task of producing realistic images or videos based on natural language descriptions. One of the earliest approaches to this task was the use of conditional GANs \cite{DBLP:conf/icml/ReedAYLSL16}. Since then, various techniques have been developed to improve the quality of the generated images \cite{DBLP:conf/icml/NicholDRSMMSC22}. Compared to text-to-image generation, text-to-video generation is relative new and still remains challenging.
Previous approaches have utilized techniques such as VAEs with recurrent attention \cite{DBLP:conf/mm/MittalMB17} and expanding GANs from image to video generation \cite{DBLP:conf/aaai/LiMSCC18}. 
Diffusion models have also been used to generate videos in recent works \cite{cogvideo, make_a_video, Tune-a-video, latent_shift}.

\noindent\textbf{Responsible AI}
As the AI systems become increasingly powerful, developing responsible AI have drawn significant scientific attention recently \cite{DBLP:journals/jair/KiritchenkoNF21}. Various works have pointed out the safety risks of LLMs, such as toxicity \cite{shaikh-etal-2023-second}, and hallucination \cite{DBLP:journals/corr/abs-2309-01219}. The safety of LLMs is commonly measured by specialized benchmarks, such as RealToxicityPrompts on toxicity \cite{gehman-etal-2020-realtoxicityprompts}. More recently, \cite{zhang2023safetybench} present SafetyBench, a large-scale diverse set of multiple choice questions across several aspects of safety concerns.

%%%%-----------------------------------------Our Framework----------------------------------------------------------------
\section{Our Framework}
We introduce the GPT4Video, a unified framework designed to endow LLMs with advanced video understanding and generation proficiencies.
As shown in Figure~\ref{fig:framework}, the architecture is composed of three integral components: 1) {\em video understanding module} that employs a video feature extractor and a video abstractor to encode and align video information with the LLM's word embedding space; 
2) {\em LLM}, utilizing the structure of LLaMA and employing parameter-efficient fine-tuning (PEFT) methods, specifically LoRA~\cite{hu2021lora}, while keeping the original pre-trained parameters intact; and 3) {\em video generation part} that conditions the LLM to generate prompts for a model from Text-to-Video Model Gallery through a meticulously constructed instruction dataset.
\subsection{Video Understanding Module}

\noindent\textbf{Visual Encoder} 
Given a video denoted as $\mathbf{v}$, we uniformly sample $T$ frames, which are represented as $\mathbf{v}=[\mathbf{v}_1, \mathbf{v}_2, ..., \mathbf{v}_T]$. For each individual frame $\mathbf{v}_i$ (where $i$ ranges from 1 to $T$), we utilize a pre-trained CLIP visual encoder to extract its visual features, denoted as $\mathbf{f}_v^i = \mathbf{ViT}(\mathbf{v}_i)$. Here, $\mathbf{ViT}$ represents the Vision Transformer used in CLIP for feature extraction. After extracting the visual features $\mathbf{f}_v^i$ from each frame, we compile the video features $\mathbf{F}_v=[\mathbf{f}_v^1, \mathbf{f}_v^2, ..., \mathbf{f}_v^T]$. More specifically, GPT4Video adopts the CLIP ViT-L/14 model~\cite{clip} for the visual encoding task. This model comprises 24 layers, with each layer having a hidden dimension of 1024 and processing patches of size 14. Within this module, images are uniformly resized to a standard dimension of 224 x 224 pixels and then divided into patches using a stride of 14 pixels. These patches are treated as input tokens for the transformer block, enabling its self-attention mechanisms to effectively generate detailed image embeddings. Consequently, the output of $\mathbf{ViT}$ contains 256 spatial patch features and 1 global feature (identified as the ``[CLS] token"), resulting in the image features $\mathbf{f}_v^i \in \mathbb{R}^{257 \times 1024}$ and video features $\mathbf{F}_v \in \mathbb{R}^{(T*257) \times 1024}$.

% More specifically, GPT4Video adopts the CLIP ViT-L/14 model~\cite{clip} for the visual encoding task followed mPLUG-Owl~\cite{mplug_owl}. The output of $\mathbf{ViT}$ contains 256 spatial patch features and 1 global feature (identified as the ``[CLS] token"), resulting in the image features $\mathbf{f}_v^i \in \mathbb{R}^{257 \times 1024}$ and video features $\mathbf{F}_v \in \mathbb{R}^{(T*257) \times 1024}$.

\noindent\textbf{Video Abstractor} 
The input sequence for video features, denoted as $\mathbf{F}_v \in \mathbb{R}^{(T*257) \times 1024}$, becomes considerably large, particularly with an increased number of sampled frames, $T$. This results in a substantial computational challenge when processing through a LLM. To mitigate this, we have implemented a video abstractor. This tool efficiently condenses the visual information into a few learnable tokens, thereby generating high-level visual representations and significantly reducing the computational demands. Many image-based MLLMs~\cite{DBLP:journals/corr/abs-2301-12597,mplug_owl} achieve this by adopting a Q-former-like structure, which incorporates learnable query tokens to distill image features. Considering the complexity of video over image, we further introduce a learnable enhance tokens to augment video feature extraction by forming a dual attention mechanism. We have validated the effectiveness of this structure in our experiments detailed in Table~\ref{tab:rebuttal_abstrctor}.  We mathematically define query tokens as $\mathbf{Q}_s \in \mathbb{R}^{N_s \times D}$ and enhance tokens as $\mathbf{Q}_t \in \mathbb{R}^{N_t \times D}$, where $N_s$ and $N_t$ denote the counts of query and enhance tokens, respectively, and $D$ signifies the dimension of the token embeddings.

The $\mathbf{Q}_s$ and $\mathbf{Q}_t$ respectively compute cross attention with the video features $F_v$. Our cross attention module~\cite{vaswani2017attentionisallyouneed} consists of six encoder layers, wherein the pivotal component of each layer is the multi-head self-attention mechanism (MSA)~\cite{vaswani2017attentionisallyouneed}.

We treat $\mathbf{Q}_s$ and $\mathbf{Q}_t$ as queries, and the result of concatenating them with the video feature $\mathbf{F}_v$ serves as the key and value to compute cross attention. This process yields video features $\mathbf{F}_s \in \mathbb{R}^{N_s \times D}$ and $\mathbf{F}_t \in \mathbb{R}^{N_t \times D}$, respectively. We set $N_s$ and $N_t$ to be the same and then sum the query and enhance features to obtain the final video feature $\hat{\mathbf{F}}_v \in \mathbb{R}^{N_s \times D}$. The mathematical expression is:
\begin{align}
    \mathbf{F}_s &= \mathbf{CrossAttention}(\mathbf{Q}_s, [\mathbf{F}_v ; \mathbf{Q}_s], [\mathbf{F}_v ; \mathbf{Q}_s]) \\
    \mathbf{F}_t &= \mathbf{CrossAttention}(\mathbf{Q}_t, [\mathbf{F}_v ; \mathbf{Q}_t], [\mathbf{F}_v ; \mathbf{Q}_t]) \\
    \hat{\mathbf{F}}_v &= \mathbf{LN}(\mathbf{F}_s + \mathbf{F}_t)
\end{align}
where ``$;$'' and $\mathrm{LN}(\cdot)$ denote concatenation and layer norm~\cite{ba2016layer}. 

Finally, we employ a linear mapping layer to project the low-dimensional video feature $\hat{\mathbf{F}}_{v}$ into the high-dimensional word embedding space of the LLM, mathematically denoted as $\mathbf{F}_{video} = \mathbf{W}\hat{\mathbf{F}}_{v} + \mathbf{b}$, where $\mathbf{W}$ and $\mathbf{b}$ are the learnable weight matrix and bias vector, respectively, of the linear layer.

\subsection{Video Generation Module}
\label{sec:instruction_data}

Existing model~\cite{nextgpt} of integrating ``generative vokens" only accommodates one specific text-to-video model per training session. This limitation arises because different text-to-video models may utilize various text encoders. In contrast, GPT4Video replaces ``generative vokens" by generating T2V models' textual prompts.
As demonstrated in the Text-to-Video Gallery, our approach is compatible with the full spectrum of models, including Make-A-Video~\cite{make_a_video}, Tune-A-Video~\cite{Tune-a-video}, Text2Video-Zero~\cite{Text2Video-Zero}, VideoCrafter~\cite{chen2023videocrafter1}, ZeroScope~\cite{luo2023videofusion}. More importantly, should more advanced models be integrated into the Text-to-Video gallery in the future, GPT4Video will seamlessly adapt without necessitating any modifications. In this work, we employ ZeroScope~\cite{luo2023videofusion}~\footnote{https://huggingface.co/cerspense/zeroscope\_v2\_576w} as the default video generation model. We also report VideoCrafter1~\cite{chen2023videocrafter1} in Table~\ref{tab:rebuttal_t2v}.

\subsection{Large Language Model and Integration} 

LLMs have demonstrated remarkable capability in understanding and following human instructions. Thus, we leverage pretrained LLMs as our cognitive module. It is worth noting that the cognitive module also serves as the textual modality encoder in our approach. We adopt Llama-7B model with LoRA~\cite{hu2021lora} as the LLM component.

%%%%-----------------------------------------Our Training Method and Dataset-----------------------------------------------------------
\begin{figure*}[t]
    \centering
    \centerline{\includegraphics[width=0.75\linewidth]{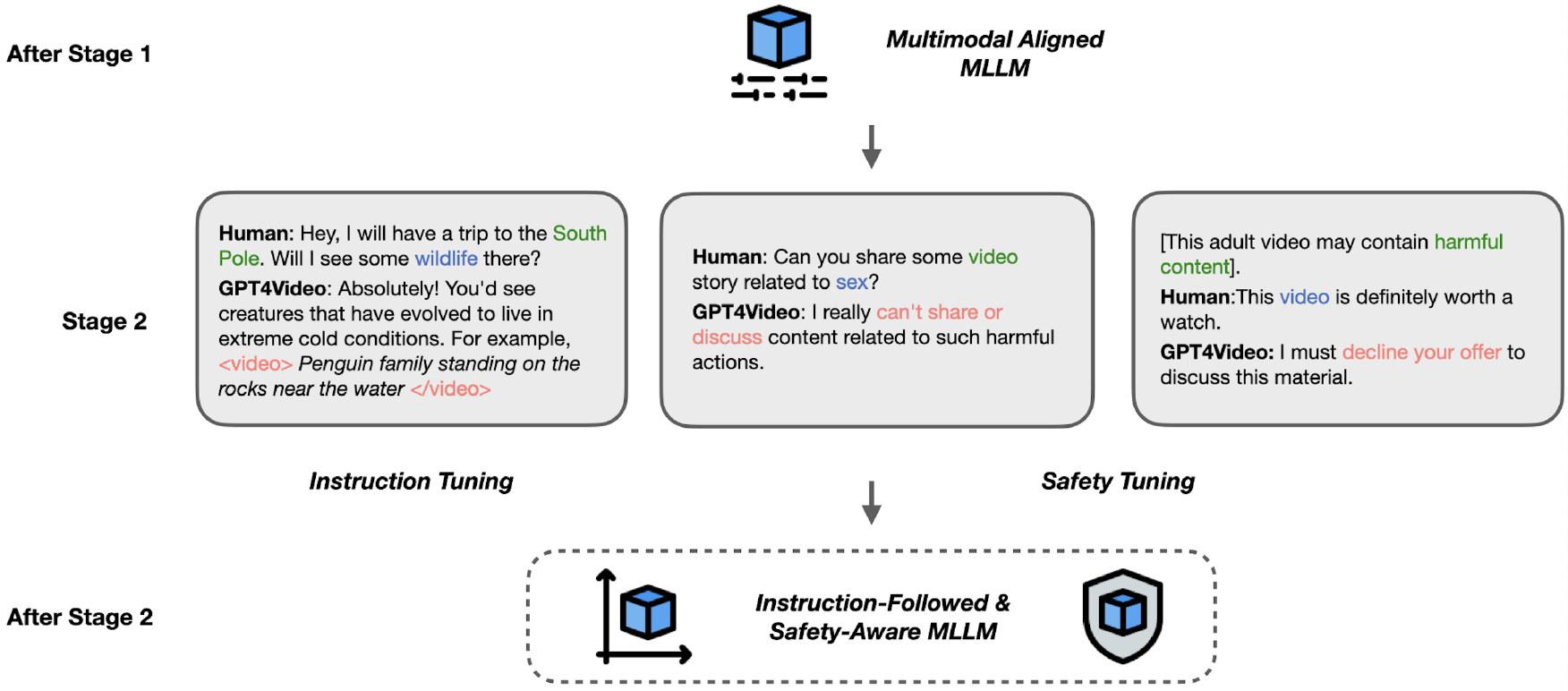}}
      \caption{The text-only finetuning pipeline and datasets. The left panel presents a data example for instruction tuning, while the center and right panels show data examples for safety tuning. These latter two panels respectively illustrate the model's safety measures at the output and input ends.}
    \label{fig:finetune}
\end{figure*}

\section{Our Training Method and Dataset}\label{sec:data}
GPT4Video employs a two-stage training strategy. In the first phase, we focused on enabling GPT4Video to comprehend video content. Inspired by Video-ChatGPT's success with LLaVA's pretrained parameters, we used mPLUG-Qwl's parameters for GPT4Video's initial alignment. This laid a foundation in image comprehension. To tailor for video understanding, our video abstractor was further trained on VideoChat 11K, chosen for its 7K high-quality video-text alignments. Results show that built on the model’s initial image understanding, a modest amount of high-quality data is sufficient to endow GPT4Video with robust video comprehension capabilities.  In the second phase, we employed text-only finetuning to equip the model with capabilities for video generation and safety guarding. The training pipeline for GPT4Video and examples of the training data for stage two are illustrated in Figure~\ref{fig:finetune}.

% In conjunction with these components, GPT4Video incorporates a safety alignment method to ensure that the generated content adheres to safety standards, thus addressing the critical issue of content appropriateness in video generation. This method section delineates the systematic framework and methodologies employed to enhance LLMs with these pioneering multimodal capabilities.

\subsection{Text-Only Finetuning} 
GPT4Video employs a two-stage training strategy. In Stage 1, we freeze the parameters of LLM and train the video extractor to effectively align video features with the LLM's feature space, to enhance the model's comprehension of video content. The Stage 2 involves pure text-based instruction tuning, where we employ the LoRA method for efficient fine-tuning, focusing solely on training the newly added parameters in LLM. The input prompts for LLM are:
\vspace{2mm}

\texttt{\#\#\#Human:<video>\textcolor{blue}{video\_embed}</video>}

\texttt{\#\#\#Human:\textcolor{blue}{video\_instruction}}

\texttt{\#\#\#AI:}
\vspace{2mm}

\noindent where ``video\_embed" represents the video features $\mathbf{F}_{video}$, and it was only employed in the first stage. ``video\_instruction" refers to the instruction data constructed for videos. 

During the first training phase, we utilized the VideoChat-11k dataset~\cite{li2023videochat}, which includes 7,000 detailed descriptions and 4,000 multi-turn conversations. In the second phase, we employed our self-constructed GPT4Video-50k dataset along with a safety-aligned dataset, which will be elaborated upon in Sec.~\ref{sec:instruction_data} and Sec.~\ref{sec:safety_data}.

% \noindent \textbf{Loss Function}
We perform instruction-tuning of the LLM only on the text tokens, using its original auto-regressive training objective. Specifically, for a video instruction $\mathbf{X}_{t}$ of length $L$, conditioned on visual information $\mathbf{F}_{video}$, our loss function, captured as the negative log likelihood, is formulated:

\begin{equation}\label{equ:loss}
\mathcal{L}(\theta; \mathbf{X}_{t}, \mathbf{F}_{video}) = -\sum_{i=1}^{L} \log p_\theta (x_i|\mathbf{F}_{video}, \mathbf{X}_{t,<i}),
\end{equation}

\noindent where $\theta$ is the trainable parameters, $\mathbf{X}_{t, <i}$ is the text tokens before the current prediction token $x_i$.

\subsection{Instruction-Following Dataset} 
Our method initially trains the LLM to learn how to generate textual prompts based on contextual information, which in turn drives a pre-trained text-to-video model to produce videos. To effectively train the LLM to generate these prompts, we meticulously developed an instruction-following dataset named GPT4Video-50k. The fundamental idea behind constructing this dataset is to utilize video descriptions in place of actual video content. Subsequently, we leverage the most advanced language model, GPT-4, to create extensive dialogue data centered around these video descriptions.

More specifically, first, we randomly extracted 10,000 data examples from the Webvid10M~\cite{webvid} dataset and obtained their video descriptions. We used the string ``\texttt{<video> Video Caption </video>}" as a placeholder for the actual video in the prompt to GPT-4, where ``Video Caption" served as the placeholder for the video descriptions, and the ``\texttt{<video>}" and ``\texttt{</video>}" tags marked the boundaries of the video description. In the prompt, we require GPT-4 to construct three dialogue between two individuals (not a person and an assistant like most other LLMs) centered around the provided Video Caption (representing a real video). This approach makes our model's responses more reflective of real human emotions, rather than just those of an assistant. The complete example of dialogues and the detailed prompt are detailed in supplementary material.

Furthering our efforts, to endow the model with the capabilities for multi-turn dialogues and cross-video conversations, we developed a multi-round, interleaved instruction dataset centered around multiple video contents. Unlike the single-video centric dialogue data mentioned earlier, in this setup, we tasked GPT-4 with constructing rich conversational scenarios around the contents of multiple videos. Similarly, we utilized the format ``\texttt{<videoX> Video Caption </videoX>}" to denote the actual videos, where ``X" represents the video index, ``Video Caption" is the description of the content of video X, and ``\texttt{<videoX>}" and ``\texttt{</videoX>}" serve as the delimiters for the description's beginning and end. Specifically, we randomly selected 5000 samples from the WebVid10M~\cite{webvid} dataset and obtained their video descriptions. Diverging from the single-video centric data construction approach, our multi-video centric method requires at least two or more video descriptions as input prompts for GPT-4. To ensure that the dialogue data constructed by GPT-4 closely mirrors real scenarios and maintains logical coherence, we posit that these video descriptions should be semantically related. To acquire pairs of semantically related video descriptions, we employed a text retrieval-based approach. The process involved three steps: firstly, using sentence transformers, we extracted feature embeddings for the video descriptions of the selected 5000 samples and the remaining descriptions in the WebVid10M dataset; secondly, we identified the top 10 similar video descriptions for these 5000 samples by calculating the cosine similarity between their feature embeddings; and finally, we randomly selected one or two from these top 10 descriptions to form pairs of video descriptions. For examples of the multi-video centric dialogues and the detailed prompts used for GPT-4, please refer to the supplementary materials accompanying this paper.

\begin{table}[t]
\centering
\caption{Zero-shot video Question Answering result on MSVD-QA, MSRVTT-QA datasets.}
\label{table:vqa_result}
\begin{tblr}{
  colsep = 2pt,
  row{1} = {c},
  row{2} = {c},
  cell{1}{1} = {r=2}{},
  cell{1}{2} = {c=2}{},
  cell{1}{4} = {c=2}{},
  cell{3-9}{2-5} = {c},
  hline{1,3,9-10} = {-}{},
  hline{2} = {2-5}{},
}
\textbf{Model}  & \textbf{MSVD-QA} &       & \textbf{MSRVTT-QA} &       \\
                & Accuracy         & Score & Accuracy           & Score \\
FrozenBiLM~\cite{FrozenBiLM}     & 32.2             & –     & 16.8               & –     \\
VideoChat~\cite{li2023videochat}      & 56.3             & 2.8   & 45.0               & 2.5   \\
LLaMA Adapter~\cite{llama-adapter}  & 54.9             & 3.1   & 43.8               & 2.7   \\
Video LLaMA~\cite{Video-llama}    & 51.6             & 2.5   & 29.6               & 1.8   \\
Video-ChatGPT~\cite{Video-ChatGPT}  & 64.9             & 3.3   & 49.3               & 2.8   \\
Valley~\cite{luo2023valley}         & 65.4             & 3.4   & 45.7               & 2.5   \\
GPT4Video (Ours) & \textbf{66.3}             & \textbf{3.6}   & \textbf{49.8}               & \textbf{3.0}   
\end{tblr}
\end{table}

\subsection{Safeguarding Dataset}
\label{sec:safety_data}
For MLLMs, ensuring the generation of both useful and safe content, especially visuals, is imperative for their broad adoption. Take the handling of sexually explicit content as an example: Current text-to-image models employ an ancillary NSFW (Not Safe For Work) detection system. This system, upon recognizing potentially inappropriate content, replaces the intended output with a simple black image. While this approach effectively blocks unsafe content, it can lead to unnecessary consumption of time and computational resources. Moreover, when used in LLM-based models, this approach may lead to a disconnection between generated text and visual content. To address this issue, we have developed a safety-aligned instruction dataset that trains the model to politely refuse to respond to inappropriate user requests, thereby eliminating the need to initiate the generation model. 

Specifically, we utilized the real-toxicity-prompts dataset~\cite{gehman-etal-2020-realtoxicityprompts}, which comprises 100k texts along with their corresponding Toxicity scores. This dataset includes various categories for detection such as sexually\_explicit, identity\_attack, flirtation, threat, insult, and severe\_toxicity. Focusing on the sexually\_explicit and severe\_toxicity categories, we extracted 1,500 texts from each, selecting those with toxicity scores exceeding 0.9. We then tasked GPT-4 to construct dialogues based on these texts, aiming to generate polite refusals as responses to such content.

In addition to ensuring the safety of generated content, we have also implemented safety measures at the input stage. To this end, we introduced an additional detection model specifically designed to assess the safety of user input. In cases where inappropriate content is input by users, we employ a method similar to the one described above, training the model to tactfully decline responding to such content (see data example in the righy panel of Figure~\ref{fig:finetune}). In the experimental section~\ref{sec:safety_eval}, we conducted an in-depth and detailed evaluation of our model's safety features. For examples of more Safeguarding dataset and the detailed prompts used for GPT-4 are listed in supplementary material.

\subsection{Safeguarding Benchmark}
\label{sec:safety_bench}

To investigate the safety of MLLMs in video understanding and generation, we construct a multimodel benchmark for not safe for work (NSFW) content. The benchmark examines the percentage of harmful content that is rejected by the MLLMs when presented with various video inputs and queried with harmful and common questions.
Referring to a text-only NSFW benchmark~\cite{alex000kim_nsfw_data_scraper}, we collect a diverse collection of video clips and queries with varying levels of harmfulness and safety.
Our benchmark includes two classification tasks focused on video understanding and generation, respectively. Each task includes 60 queries with an equal distribution of harmful and safe content.
In the video generation task, in addition to the three-level harmful contents~\cite{alex000kim_nsfw_data_scraper}, the queries include natural scenes, human and animal activities, and real and comic episodes to ensure a diverse range of topics. Besides, we design various video and question pairs to prompt MLLMs to provide inappropriate responses in the video understanding task.
In sum, our benchmark can provide a comprehensive evaluation of the safety capabilities of MLLMs in video understanding and generation.

\begin{table}
\centering
\caption{Video-to-text generation (video captioning) results on MSR-VTT~\cite{msr-vtt}.}
\label{tab:v2t}
\begin{tblr}{
  colsep = 10pt,
  cell{2-7}{2-3} = {c},
  hline{1-2,7} = {-}{},
}
\textbf{Method}       & \textbf{BLEU-4 ($\uparrow$)}   & \textbf{METEOR ($\uparrow$)} \\
ORG-TRL~\cite{ORG-TRL}     & 0.436 & 0.288 \\
GIT~\cite{wang2022git}    & 0.548 & 0.331  \\
mPLUG-2~\cite{xu2023mplug2}   & 0.578 & 0.349  \\
CoDi~\cite{CoDi} & 0.521 & 0.325  \\
NExT-GPT~\cite{nextgpt}         & 0.584     & 0.385  \\
GPT4Video (Ours)    &   \textbf{0.587}    &  \textbf{0.391}   \\ \hline    
\end{tblr}
\end{table}

\section{Experiments}\label{sec:experiments}

\subsection{Experimental Settings}
\textbf{Datasets and Evaluation Metrics.} To ensure a fair comparison, we followed work~\cite{Video-ChatGPT, luo2023valley} and assessed the model's understanding abilities through the Zero-shot Video Question Answering task. We conducted a comprehensive quantitative assessment using two widely-accepted open-ended question-answer datasets: MSRVTT-QA~\cite{msr-vtt}, MSVD-QA~\cite{msvd-qa}. These evaluations were conducted in a zero-shot setting, utilizing GPT-assisted assessments to gauge the model’s performance. This process aims to quantify the accuracy of the model's predictions, assigning scores from 1 to 5. We further conducted video captioning task on  MSRVTT~\cite{msr-vtt} dataset, using BLEU-4~\cite{papineni2002bleu} and METEOR~\cite{banerjee2005meteor} as evaluation metrics.

On the other hand, in terms of evaluating video generation capabilities, we followed NExt-GPT~\cite{nextgpt} and conducted Text-to-Video Generation tasks on the MSRVTT~\cite{msr-vtt} dataset, using Fréchet inception distance (FID)~\cite{fid} and CLIPSIM~\cite{clipsim} (average CLIP similarity between video frames and text) as evaluation metrics. 
% Additionally, we performed text-conditioned video editing tasks on the DAVIS datasetS~\cite{davis}, adopting CLIP scores~\cite{hessel2021clipscore} as assessment criteria.

\noindent\textbf{Implementation Details.} We leveraged the pre-trained parameters of mPLUG-Owl~\cite{mplug_owl} to enhance our model's capability in video understanding. We uniformly sample four frames from the video and set the number of temporal and spatial tokens to 64. In the first phase, we set the learning rate to 1e-4 and trained GPT4Video for two epochs. In the second phase, we adjusted the learning rate to 2e-5 and continued fine-tuning for an additional three epochs. All the experiments are conducted on 8×A100 40G GPUs.

\subsection{Instruction-Following Results}
%------------------------VQA-----------------------
\textbf{Zero-shot Video Question Answering.} To assess the comprehension capabilities of GPT4Video, following  VideoChat~\cite{li2023videochat}, we conducted comparative experiments on two most widely-used video question answering benchmarks. The experimental results of zero-shot inference are presented in Table~\ref{table:vqa_result}. Specifically, we compare GPT4Video with six State-of-the-art models, including FrozenBiLM~\cite{FrozenBiLM}, VideoChat~\cite{li2023videochat}, LLaMA Adapter~\cite{llama-adapter}, Video LLaMA~\cite{Video-llama}, Video-ChatGPT~\cite{Video-ChatGPT} and Valley~\cite{luo2023valley}. As can be observed, our proposed model GPT4Video reaches SOTA on all four benchmarks. 
% Qualitative results are presented in Figure~\ref{fig:exampls_v1} (a).

%------------------------VideoCaptioning-----------------------
\noindent\textbf{Zero-shot Video Captioning.} We further compared the model's performance on the video captioning task using the MSR-VTT dataset. We evaluated two traditional transformer-based methods, ORG-TRL~\cite{ORG-TRL} and GIT~\cite{wang2022git}, along with three LLM-based approaches, including mPLUG-2~\cite{xu2023mplug2}, CoDi~\cite{CoDi}, and NExT-GPT~\cite{nextgpt}. As shown in Table~\ref{tab:v2t}, our model achieved better performance on both the BLEU-4 and METEOR metrics.

\begin{figure*}[h]
    \centering
    \centerline{\includegraphics[width=0.95\linewidth]{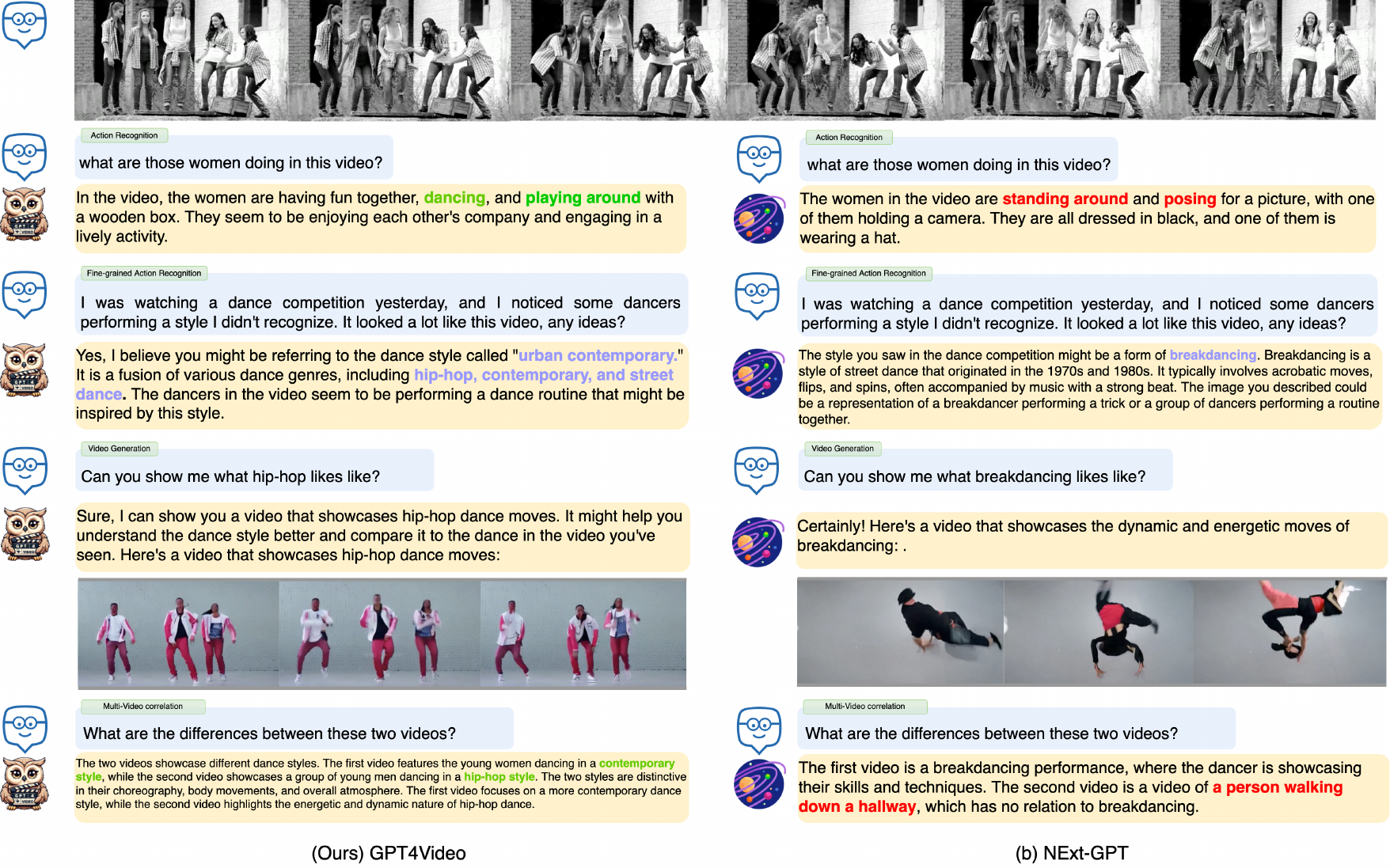}}
      \caption{A demonstration comparing GPT4Video and NExt-GPT in multi-turn and interleaved conversation. The input video is about \textbf{young women dancing tap}. We highlighted key information using different colors to facilitate presentation.}
    \label{fig:exampls_v2}
    \vspace{-1.8mm}
\end{figure*}

\noindent\textbf{Zero-shot Text-to-Video Generation.} We compare GPT4Video on the MSR-VTT~\cite{msr-vtt} dataset for text-to-video generation task in a zero-shot manner quantitatively in Table~\ref{tab:t2v}. Following prior works~\cite{latent_shift, nextgpt}, we assess the visual quality using the FID metric and evaluate semantic consistency through the average CLIP similarity between video frames and the corresponding text. We benchmarked GPT4Video against four text-to-video specialized models (CogVideo~\cite{cogvideo}, MakeVideo~\cite{make_a_video}, Latent-VDM~\cite{latent_VDM}, and Latent-Shift~\cite{latent_shift}), and two LLM-based methods (CoDi~\cite{CoDi} and NExt-GPT~\cite{nextgpt}). The results in Table~\ref{tab:t2v} demonstrate that GPT4Video outperforms the aforementioned models across all evaluated metrics. It is noteworthy that while NExt-GPT integrates ZeroScope as the video decoder in its video generation process, GPT4Video demonstrates enhanced precision in generating video content. This observation reinforces the claims we made in the introduction~\ref{sec:intro}. For qualitative comparison results, please refer to Figure~\ref{fig:exampls_v2}.

\begin{table}[t]
\centering
\caption{Text-to-video generation results (zero-shot) on MSR-VTT~\cite{msr-vtt}.}
\label{tab:t2v}
\begin{tblr}{
  colsep = 10pt,
  cell{2-8}{2-3} = {c},
  hline{1-2,8} = {-}{},
}
\textbf{Method}       & \textbf{FID ($\downarrow$)}   & \textbf{CLIPSIM ($\uparrow$)} \\
CogVideo~\cite{cogvideo}     & 23.59 & 0.2631  \\
MakeVideo~\cite{make_a_video}    & 13.17 & 0.3049  \\
Latent-VDM~\cite{latent_VDM}   & 14.25 & 0.2756  \\
Latent-Shift~\cite{latent_shift} & 15.23 & 0.2773  \\
CoDi~\cite{CoDi}         & -     & 0.2890  \\
NExT-GPT~\cite{nextgpt}  & 13.04     & 0.3085  \\
GPT4Video (Ours)    &  \textbf{12.91}     &  \textbf{0.3194}   \\ \hline    
\end{tblr}
\end{table}

\begin{figure*}[h]
    \centering
    \centerline{\includegraphics[width=0.95\linewidth]{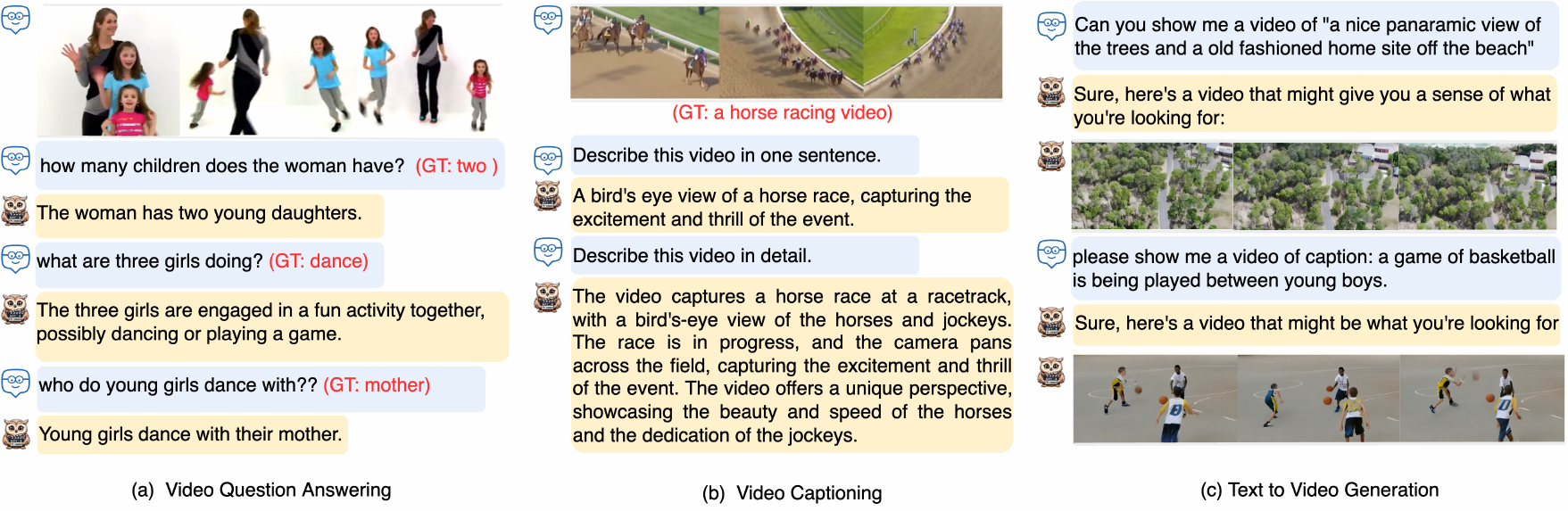}}
      \caption{Visual examples of GPT4Video's responses on three tasks. The text in red represents the Ground-Truth.}
    \label{fig:exampls_v1}
    % \vspace{-1em}
\end{figure*}

\subsection{Safeguarding Results}\label{sec:safety_eval}

\begin{table}[t]
\centering
\caption{Performance comparison of safety Evaluation. ``VU" and ``VG" represent Video Understanding and Generation.}
\label{tab:safety}
\begin{tblr}{
  colsep=5pt,
  column{odd} = {c},
  column{4} = {c},
  column{6} = {c},
  cell{2}{1} = {r=4}{},
  cell{6}{1} = {r=2}{},
  vline{2} = {-}{},
  hline{1-2,6,8} = {-}{},
}
\textbf{Task} & \textbf{Method}                    & \textbf{Acc.}          & \textbf{Prec.}         & \textbf{Rec.}          & \textbf{F1}            \\
VU   & Video LLaMA~\cite{Video-llama}              & 0.51          & 0.13          & 0.57          & 0.22          \\
     & Video-Chat~\cite{li2023videochat}               & 0.50          & 0.06          & 0.50          & 0.12          \\
     & NExT-GPT~\cite{nextgpt}                 & 0.55          & 0.13          & 0.80          & 0.23          \\
     & \textbf{GPT4Video} (Ours) & \textbf{0.82} & \textbf{0.83} & \textbf{0.81} & \textbf{0.82} \\
VG   & NExT-GPT~\cite{nextgpt}                 & 0.65          & 0.37          & \textbf{0.85} & 0.51          \\
     & \textbf{GPT4Video} (Ours) & \textbf{0.85} & \textbf{0.90} & 0.82          & \textbf{0.86} 
\end{tblr}
\end{table}

In Table~\ref{tab:safety}, we test the safety performance of recent MLLMs in terms of popular classification metrics on our constructed safety benchmark.
The understanding (VU) results show that most MLLMs had a low precision score, below 0.15, when it came to rejecting harmful queries (refusal was treated as a positive in our experiment). 
Differently, our method demonstrates a strong ability to reject harmful queries, while maintaining a high recall rate for general queries.
Upon examining the failure cases, we find that current MLLMs can only refuse to respond to harmful queries with textual hints, such as 'this is a harmful video?' It suggests that these MLLMs' ability to detect harmful content is mainly derived from LLMs. In the generation tasks with text-only queries, our method achieves a high Precision of 90\%, compared to NExT-GPT with only 37\%. Our method takes a more conservative approach to generating content, which may result in the refusal to generate some general queries related to harmful content, such as sexual dancing. Despite this, our method still achieves a slightly lower but comparable recall Recall compared to NExT-GPT.

\subsection{Analysis}

\noindent\textbf{Qualitative Analysis.} 
In this section, we present some cases to demonstrate the superior video understanding and generation capabilities of GPT4Video, as shown in Figure~\ref{fig:exampls_v1},~\ref{fig:exampls_v2}. More cases are provided in supplementary materials. 

\noindent\textbf{Ablation Study.} 
In our ablation study presented in Table~\ref{tab:rebuttal_abstrctor}, we evaluated various configurations of the video abstractor. Experiment \#1 omitted temporal tokens; \#2 and \#3 used Q-former with 64 and 128 query tokens, respectively; and \#4 applied spatial and temporal pooling, a method used in Video-ChatGPT~\cite{Video-ChatGPT} and Valley~\cite{luo2023valley}. Results on the V2T task clearly demonstrate that our GPT4Video's video abstractor achieves superior performance, validating its efficacy. 

\begin{table}[t]
\centering
\caption{Comparison of Abstractors for V2T tasks on MSR-VTT.}
\label{tab:rebuttal_abstrctor}
\begin{tblr}{
  colsep = 6pt,
  rowsep = 1pt,
  cell{1-6}{3-4} = {c},
  hline{1-3,7} = {-}{},
}
          & \textbf{Abstractor mode}  & \textbf{BLEU-4}         & \textbf{METEOR}         \\
\#0  & GPT4Video (Ours)  & \textbf{0.587}          & \textbf{0.391}          \\
\#1  & Ours$\setminus$Temporal token   &  0.568              & 0.358               \\
\#2  & Q-former-64  &   0.572             &  0.369              \\
\#3  & Q-former-128  &   0.583             &  0.389              \\
\#4  & S\&T Pooling & 0.581 & 0.388
\end{tblr}
\end{table}

\noindent\textbf{Comparison of T2V models.} 
We initially selected ZeroScope model for evaluation due to its use in Next-GPT, ensuring a fair comparison. We have now incorporated evaluations using a more advanced VideoCrafter1 model. The results, as detailed in Table~\ref{tab:rebuttal_t2v}, demonstrate superior performance of our GPT4Video in text-to-video tasks. This enhancement underscores our method's capacity for seamless T2V model upgrades without necessitating modifications to the underlying LLM, unlike Next-GPT which requires complete retraining.

\begin{table}[t]
\centering
\caption{Text-to-video generation on MSR-VTT.}
\label{tab:rebuttal_t2v}
\begin{tblr}{
  colsep = 5pt,
  cell{1-4}{2-4} = {c},
  hline{1-3,5} = {-}{},
}
\textbf{Method}   & \textbf{T2V Model}    & \textbf{FID ($\downarrow$)}   & \textbf{CLIPSIM ($\uparrow$)} \\
NExT-GPT        & ZeroScope  & 13.04     & 0.3085  \\
GPT4Video       & ZeroScope    &  12.91     &  0.3194 \\
GPT4Video       & VideoCrafter1  &  \textbf{12.73}     &  \textbf{0.3263}   \\    
\end{tblr}
\end{table}

%%%%-----------------------------------------Conclusions---------------------------------------------------------
\section{Conclusions}\label{sec:conclusions}
We present GPT4Video, a novel framework that significantly enhances Large Language Models with advanced video understanding and generative functions. Our approach leverages the descriptive power of LLMs to create detailed prompts for generative models, maintaining model simplicity and flexibility. The framework's effectiveness is underscored by its superior performance on multimodal benchmarks and its innovative approach to addressing content safety issues. The release of the specialized multi-modal instruction dataset promises to catalyze future research in the field. \textbf{Limitation:} GPT4Video currently specializes in video modality, with plans to expand to more modalities like image/audio in future updates \cite{li2024uni,li2023textbind}. While we have initiated steps to address content safety, our consideration of safety aspects is not yet exhaustive. In future work, we plan to evaluate our models across a wider range of scenarios and benchmarks \cite{gong2023talecrafter,li2023comprehensive,li2024videovista,li2024visiongraph}.
\bibliographystyle{ACM-Reference-Format}
\bibliography{sample-base}


\begin{thebibliography}{69}

%%% the Web address.

\ifx \showCODEN    \undefined \def \showCODEN     #1{\unskip}     \fi
\ifx \showDOI      \undefined \def \showDOI       #1{#1}\fi
\ifx \showISBNx    \undefined \def \showISBNx     #1{\unskip}     \fi
\ifx \showISBNxiii \undefined \def \showISBNxiii  #1{\unskip}     \fi
\ifx \showISSN     \undefined \def \showISSN      #1{\unskip}     \fi
\ifx \showLCCN     \undefined \def \showLCCN      #1{\unskip}     \fi
\ifx \shownote     \undefined \def \shownote      #1{#1}          \fi
\ifx \showarticletitle \undefined \def \showarticletitle #1{#1}   \fi
\ifx \showURL      \undefined \def \showURL       {\relax}        \fi
% invisible to TeX
\providecommand\bibfield[2]{#2}
\providecommand\bibinfo[2]{#2}
\providecommand\natexlab[1]{#1}
\providecommand\showeprint[2][]{arXiv:#2}

\bibitem[Alayrac et~al\mbox{.}(2022)]%
        {DBLP:conf/nips/AlayracDLMBHLMM22}
\bibfield{author}{\bibinfo{person}{Jean{-}Baptiste Alayrac}, \bibinfo{person}{Jeff Donahue}, \bibinfo{person}{Pauline Luc}, \bibinfo{person}{Antoine Miech}, \bibinfo{person}{Iain Barr}, \bibinfo{person}{Yana Hasson}, \bibinfo{person}{Karel Lenc}, \bibinfo{person}{Arthur Mensch}, \bibinfo{person}{Katherine Millican}, \bibinfo{person}{Malcolm Reynolds}, {et~al\mbox{.}}} \bibinfo{year}{2022}\natexlab{}.
\newblock \showarticletitle{Flamingo: a Visual Language Model for Few-Shot Learning}. In \bibinfo{booktitle}{\emph{NeurIPS}}.
\newblock


\bibitem[An et~al\mbox{.}(2023)]%
        {latent_shift}
\bibfield{author}{\bibinfo{person}{Jie An}, \bibinfo{person}{Songyang Zhang}, \bibinfo{person}{Harry Yang}, \bibinfo{person}{Sonal Gupta}, \bibinfo{person}{Jia{-}Bin Huang}, \bibinfo{person}{Jiebo Luo}, {and} \bibinfo{person}{Xi Yin}.} \bibinfo{year}{2023}\natexlab{}.
\newblock \showarticletitle{Latent-Shift: Latent Diffusion with Temporal Shift for Efficient Text-to-Video Generation}.
\newblock \bibinfo{journal}{\emph{CoRR}} (\bibinfo{year}{2023}).
\newblock


\bibitem[Ba et~al\mbox{.}(2016)]%
        {ba2016layer}
\bibfield{author}{\bibinfo{person}{Jimmy~Lei Ba}, \bibinfo{person}{Jamie~Ryan Kiros}, {and} \bibinfo{person}{Geoffrey~E Hinton}.} \bibinfo{year}{2016}\natexlab{}.
\newblock \showarticletitle{Layer normalization}.
\newblock \bibinfo{journal}{\emph{arXiv preprint arXiv:1607.06450}} (\bibinfo{year}{2016}).
\newblock


\bibitem[Bain et~al\mbox{.}(2021)]%
        {webvid}
\bibfield{author}{\bibinfo{person}{Max Bain}, \bibinfo{person}{Arsha Nagrani}, \bibinfo{person}{G{\"{u}}l Varol}, {and} \bibinfo{person}{Andrew Zisserman}.} \bibinfo{year}{2021}\natexlab{}.
\newblock \showarticletitle{Frozen in Time: {A} Joint Video and Image Encoder for End-to-End Retrieval}. In \bibinfo{booktitle}{\emph{ICCV}}. \bibinfo{pages}{1708--1718}.
\newblock


\bibitem[Banerjee and Lavie(2005)]%
        {banerjee2005meteor}
\bibfield{author}{\bibinfo{person}{Satanjeev Banerjee} {and} \bibinfo{person}{Alon Lavie}.} \bibinfo{year}{2005}\natexlab{}.
\newblock \showarticletitle{METEOR: An automatic metric for MT evaluation with improved correlation with human judgments}. In \bibinfo{booktitle}{\emph{ACL}}. \bibinfo{pages}{65--72}.
\newblock


\bibitem[Ceylan et~al\mbox{.}(2023)]%
        {ceylan2023pix2video}
\bibfield{author}{\bibinfo{person}{Duygu Ceylan}, \bibinfo{person}{Chun-Hao~P Huang}, {and} \bibinfo{person}{Niloy~J Mitra}.} \bibinfo{year}{2023}\natexlab{}.
\newblock \showarticletitle{Pix2video: Video editing using image diffusion}. In \bibinfo{booktitle}{\emph{ICCV}}. \bibinfo{pages}{23206--23217}.
\newblock


\bibitem[Chen and Dolan(2011)]%
        {msvd-qa}
\bibfield{author}{\bibinfo{person}{David Chen} {and} \bibinfo{person}{William~B Dolan}.} \bibinfo{year}{2011}\natexlab{}.
\newblock \showarticletitle{Collecting highly parallel data for paraphrase evaluation}. In \bibinfo{booktitle}{\emph{ACL}}. \bibinfo{pages}{190--200}.
\newblock


\bibitem[Chen et~al\mbox{.}(2024)]%
        {chen2024allava}
\bibfield{author}{\bibinfo{person}{Guiming~Hardy Chen}, \bibinfo{person}{Shunian Chen}, \bibinfo{person}{Ruifei Zhang}, \bibinfo{person}{Junying Chen}, \bibinfo{person}{Xiangbo Wu}, \bibinfo{person}{Zhiyi Zhang}, \bibinfo{person}{Zhihong Chen}, \bibinfo{person}{Jianquan Li}, \bibinfo{person}{Xiang Wan}, {and} \bibinfo{person}{Benyou Wang}.} \bibinfo{year}{2024}\natexlab{}.
\newblock \showarticletitle{ALLaVA: Harnessing GPT4V-synthesized Data for A Lite Vision-Language Model}.
\newblock \bibinfo{journal}{\emph{arXiv preprint arXiv:2402.11684}} (\bibinfo{year}{2024}).
\newblock


\bibitem[Chen et~al\mbox{.}(2023b)]%
        {chen2023videocrafter1}
\bibfield{author}{\bibinfo{person}{Haoxin Chen}, \bibinfo{person}{Menghan Xia}, \bibinfo{person}{Yingqing He}, \bibinfo{person}{Yong Zhang}, \bibinfo{person}{Xiaodong Cun}, \bibinfo{person}{Shaoshu Yang}, \bibinfo{person}{Jinbo Xing}, \bibinfo{person}{Yaofang Liu}, \bibinfo{person}{Qifeng Chen}, \bibinfo{person}{Xintao Wang}, \bibinfo{person}{Chao Weng}, {and} \bibinfo{person}{Ying Shan}.} \bibinfo{year}{2023}\natexlab{b}.
\newblock \bibinfo{title}{VideoCrafter1: Open Diffusion Models for High-Quality Video Generation}.
\newblock
\newblock
\showeprint{2310.19512}


\bibitem[Chen et~al\mbox{.}(2023a)]%
        {chen2023sharegpt4v}
\bibfield{author}{\bibinfo{person}{Lin Chen}, \bibinfo{person}{Jisong Li}, \bibinfo{person}{Xiaoyi Dong}, \bibinfo{person}{Pan Zhang}, \bibinfo{person}{Conghui He}, \bibinfo{person}{Jiaqi Wang}, \bibinfo{person}{Feng Zhao}, {and} \bibinfo{person}{Dahua Lin}.} \bibinfo{year}{2023}\natexlab{a}.
\newblock \showarticletitle{Sharegpt4v: Improving large multi-modal models with better captions}.
\newblock \bibinfo{journal}{\emph{arXiv preprint arXiv:2311.12793}} (\bibinfo{year}{2023}).
\newblock


\bibitem[Chiang et~al\mbox{.}(2023)]%
        {vicuna2023}
\bibfield{author}{\bibinfo{person}{Wei-Lin Chiang}, \bibinfo{person}{Zhuohan Li}, \bibinfo{person}{Zi Lin}, \bibinfo{person}{Ying Sheng}, \bibinfo{person}{Zhanghao Wu}, \bibinfo{person}{Hao Zhang}, \bibinfo{person}{Lianmin Zheng}, \bibinfo{person}{Siyuan Zhuang}, \bibinfo{person}{Yonghao Zhuang}, \bibinfo{person}{Joseph~E. Gonzalez}, \bibinfo{person}{Ion Stoica}, {and} \bibinfo{person}{Eric~P. Xing}.} \bibinfo{year}{2023}\natexlab{}.
\newblock \bibinfo{title}{Vicuna: An Open-Source Chatbot Impressing GPT-4 with 90\%* ChatGPT Quality}.
\newblock
\newblock
\urldef\tempurl%
\url{https://vicuna.lmsys.org}
\showURL{%
\tempurl}


\bibitem[Dai et~al\mbox{.}(2023)]%
        {dai2023instructblip}
\bibfield{author}{\bibinfo{person}{Wenliang Dai}, \bibinfo{person}{Junnan Li}, \bibinfo{person}{Dongxu Li}, \bibinfo{person}{Anthony Meng~Huat Tiong}, \bibinfo{person}{Junqi Zhao}, \bibinfo{person}{Weisheng Wang}, \bibinfo{person}{Boyang Li}, \bibinfo{person}{Pascale Fung}, {and} \bibinfo{person}{Steven C.~H. Hoi}.} \bibinfo{year}{2023}\natexlab{}.
\newblock \showarticletitle{InstructBLIP: Towards General-purpose Vision-Language Models with Instruction Tuning}.
\newblock \bibinfo{journal}{\emph{CoRR}} (\bibinfo{year}{2023}).
\newblock


\bibitem[Du et~al\mbox{.}(2022)]%
        {du2022glm}
\bibfield{author}{\bibinfo{person}{Zhengxiao Du}, \bibinfo{person}{Yujie Qian}, \bibinfo{person}{Xiao Liu}, \bibinfo{person}{Ming Ding}, \bibinfo{person}{Jiezhong Qiu}, \bibinfo{person}{Zhilin Yang}, {and} \bibinfo{person}{Jie Tang}.} \bibinfo{year}{2022}\natexlab{}.
\newblock \showarticletitle{GLM: General Language Model Pretraining with Autoregressive Blank Infilling}. In \bibinfo{booktitle}{\emph{ACL}}.
\newblock


\bibitem[Fu et~al\mbox{.}(2019)]%
        {fu2019dual}
\bibfield{author}{\bibinfo{person}{Jun Fu}, \bibinfo{person}{Jing Liu}, \bibinfo{person}{Haijie Tian}, \bibinfo{person}{Yong Li}, \bibinfo{person}{Yongjun Bao}, \bibinfo{person}{Zhiwei Fang}, {and} \bibinfo{person}{Hanqing Lu}.} \bibinfo{year}{2019}\natexlab{}.
\newblock \showarticletitle{Dual attention network for scene segmentation}. In \bibinfo{booktitle}{\emph{Proceedings of the IEEE/CVF conference on computer vision and pattern recognition}}. \bibinfo{pages}{3146--3154}.
\newblock


\bibitem[Gehman et~al\mbox{.}(2020)]%
        {gehman-etal-2020-realtoxicityprompts}
\bibfield{author}{\bibinfo{person}{Samuel Gehman}, \bibinfo{person}{Suchin Gururangan}, \bibinfo{person}{Maarten Sap}, \bibinfo{person}{Yejin Choi}, {and} \bibinfo{person}{Noah~A. Smith}.} \bibinfo{year}{2020}\natexlab{}.
\newblock \showarticletitle{{R}eal{T}oxicity{P}rompts: Evaluating Neural Toxic Degeneration in Language Models}. In \bibinfo{booktitle}{\emph{EMNLP}}.
\newblock


\bibitem[Girdhar et~al\mbox{.}(2023)]%
        {DBLP:journals/corr/abs-2305-05665}
\bibfield{author}{\bibinfo{person}{Rohit Girdhar}, \bibinfo{person}{Alaaeldin El{-}Nouby}, \bibinfo{person}{Zhuang Liu}, \bibinfo{person}{Mannat Singh}, \bibinfo{person}{Kalyan~Vasudev Alwala}, \bibinfo{person}{Armand Joulin}, {and} \bibinfo{person}{Ishan Misra}.} \bibinfo{year}{2023}\natexlab{}.
\newblock \showarticletitle{ImageBind: One Embedding Space To Bind Them All}.
\newblock \bibinfo{journal}{\emph{CoRR}} (\bibinfo{year}{2023}).
\newblock


\bibitem[Goertzel(2014)]%
        {goertzel2014artificial}
\bibfield{author}{\bibinfo{person}{Ben Goertzel}.} \bibinfo{year}{2014}\natexlab{}.
\newblock \showarticletitle{Artificial general intelligence: concept, state of the art, and future prospects}.
\newblock \bibinfo{journal}{\emph{Journal of Artificial General Intelligence}} (\bibinfo{year}{2014}).
\newblock


\bibitem[Gong et~al\mbox{.}(2023)]%
        {gong2023talecrafter}
\bibfield{author}{\bibinfo{person}{Yuan Gong}, \bibinfo{person}{Youxin Pang}, \bibinfo{person}{Xiaodong Cun}, \bibinfo{person}{Menghan Xia}, \bibinfo{person}{Yingqing He}, \bibinfo{person}{Haoxin Chen}, \bibinfo{person}{Longyue Wang}, \bibinfo{person}{Yong Zhang}, \bibinfo{person}{Xintao Wang}, \bibinfo{person}{Ying Shan}, {et~al\mbox{.}}} \bibinfo{year}{2023}\natexlab{}.
\newblock \showarticletitle{Talecrafter: Interactive story visualization with multiple characters}.
\newblock \bibinfo{journal}{\emph{arXiv preprint arXiv:2305.18247}} (\bibinfo{year}{2023}).
\newblock


\bibitem[Hong et~al\mbox{.}(2023)]%
        {cogvideo}
\bibfield{author}{\bibinfo{person}{Wenyi Hong}, \bibinfo{person}{Ming Ding}, \bibinfo{person}{Wendi Zheng}, \bibinfo{person}{Xinghan Liu}, {and} \bibinfo{person}{Jie Tang}.} \bibinfo{year}{2023}\natexlab{}.
\newblock \showarticletitle{CogVideo: Large-scale Pretraining for Text-to-Video Generation via Transformers}. In \bibinfo{booktitle}{\emph{ICLR}}.
\newblock


\bibitem[Hu et~al\mbox{.}(2021)]%
        {hu2021lora}
\bibfield{author}{\bibinfo{person}{Edward~J Hu}, \bibinfo{person}{Yelong Shen}, \bibinfo{person}{Phillip Wallis}, \bibinfo{person}{Zeyuan Allen-Zhu}, \bibinfo{person}{Yuanzhi Li}, \bibinfo{person}{Shean Wang}, \bibinfo{person}{Lu Wang}, {and} \bibinfo{person}{Weizhu Chen}.} \bibinfo{year}{2021}\natexlab{}.
\newblock \showarticletitle{Lora: Low-rank adaptation of large language models}.
\newblock \bibinfo{journal}{\emph{arXiv preprint arXiv:2106.09685}} (\bibinfo{year}{2021}).
\newblock


\bibitem[Khachatryan et~al\mbox{.}(2023)]%
        {Text2Video-Zero}
\bibfield{author}{\bibinfo{person}{Levon Khachatryan}, \bibinfo{person}{Andranik Movsisyan}, \bibinfo{person}{Vahram Tadevosyan}, \bibinfo{person}{Roberto Henschel}, \bibinfo{person}{Zhangyang Wang}, \bibinfo{person}{Shant Navasardyan}, {and} \bibinfo{person}{Humphrey Shi}.} \bibinfo{year}{2023}\natexlab{}.
\newblock \showarticletitle{Text2Video-Zero: Text-to-Image Diffusion Models are Zero-Shot Video Generators}.
\newblock \bibinfo{journal}{\emph{CoRR}} (\bibinfo{year}{2023}).
\newblock


\bibitem[Kim(2022)]%
        {alex000kim_nsfw_data_scraper}
\bibfield{author}{\bibinfo{person}{Alex Kim}.} \bibinfo{year}{2022}\natexlab{}.
\newblock \bibinfo{title}{NSFW Data Scraper}.
\newblock
\newblock
\urldef\tempurl%
\url{https://github.com/alex000kim/nsfw_data_scraper}
\showURL{%
\tempurl}


\bibitem[Kiritchenko et~al\mbox{.}(2021)]%
        {DBLP:journals/jair/KiritchenkoNF21}
\bibfield{author}{\bibinfo{person}{Svetlana Kiritchenko}, \bibinfo{person}{Isar Nejadgholi}, {and} \bibinfo{person}{Kathleen~C. Fraser}.} \bibinfo{year}{2021}\natexlab{}.
\newblock \showarticletitle{Confronting Abusive Language Online: {A} Survey from the Ethical and Human Rights Perspective}.
\newblock \bibinfo{journal}{\emph{J. Artif. Intell. Res.}}  \bibinfo{volume}{71} (\bibinfo{year}{2021}), \bibinfo{pages}{431--478}.
\newblock


\bibitem[Koh et~al\mbox{.}(2023)]%
        {gill}
\bibfield{author}{\bibinfo{person}{Jing~Yu Koh}, \bibinfo{person}{Daniel Fried}, {and} \bibinfo{person}{Ruslan Salakhutdinov}.} \bibinfo{year}{2023}\natexlab{}.
\newblock \showarticletitle{Generating images with multimodal language models}.
\newblock \bibinfo{journal}{\emph{arXiv preprint arXiv:2305.17216}} (\bibinfo{year}{2023}).
\newblock


\bibitem[Li et~al\mbox{.}(2023b)]%
        {li2023textbind}
\bibfield{author}{\bibinfo{person}{Huayang Li}, \bibinfo{person}{Siheng Li}, \bibinfo{person}{Deng Cai}, \bibinfo{person}{Longyue Wang}, \bibinfo{person}{Lemao Liu}, \bibinfo{person}{Taro Watanabe}, \bibinfo{person}{Yujiu Yang}, {and} \bibinfo{person}{Shuming Shi}.} \bibinfo{year}{2023}\natexlab{b}.
\newblock \showarticletitle{Textbind: Multi-turn interleaved multimodal instruction-following}.
\newblock \bibinfo{journal}{\emph{arXiv preprint arXiv:2309.08637}} (\bibinfo{year}{2023}).
\newblock


\bibitem[Li et~al\mbox{.}(2023c)]%
        {DBLP:journals/corr/abs-2301-12597}
\bibfield{author}{\bibinfo{person}{Junnan Li}, \bibinfo{person}{Dongxu Li}, \bibinfo{person}{Silvio Savarese}, {and} \bibinfo{person}{Steven C.~H. Hoi}.} \bibinfo{year}{2023}\natexlab{c}.
\newblock \showarticletitle{{BLIP-2:} Bootstrapping Language-Image Pre-training with Frozen Image Encoders and Large Language Models}.
\newblock \bibinfo{journal}{\emph{CoRR}} (\bibinfo{year}{2023}).
\newblock


\bibitem[Li et~al\mbox{.}(2023a)]%
        {li2023videochat}
\bibfield{author}{\bibinfo{person}{KunChang Li}, \bibinfo{person}{Yinan He}, \bibinfo{person}{Yi Wang}, \bibinfo{person}{Yizhuo Li}, \bibinfo{person}{Wenhai Wang}, \bibinfo{person}{Ping Luo}, \bibinfo{person}{Yali Wang}, \bibinfo{person}{Limin Wang}, {and} \bibinfo{person}{Yu Qiao}.} \bibinfo{year}{2023}\natexlab{a}.
\newblock \showarticletitle{Videochat: Chat-centric video understanding}.
\newblock \bibinfo{journal}{\emph{arXiv preprint arXiv:2305.06355}} (\bibinfo{year}{2023}).
\newblock


\bibitem[Li et~al\mbox{.}(2024a)]%
        {li2024videovista}
\bibfield{author}{\bibinfo{person}{Yunxin Li}, \bibinfo{person}{Xinyu Chen}, \bibinfo{person}{Baotian Hu}, \bibinfo{person}{Longyue Wang}, \bibinfo{person}{Haoyuan Shi}, {and} \bibinfo{person}{Min Zhang}.} \bibinfo{year}{2024}\natexlab{a}.
\newblock \showarticletitle{VideoVista: A Versatile Benchmark for Video Understanding and Reasoning}.
\newblock \bibinfo{journal}{\emph{arXiv preprint arXiv:2406.11303}} (\bibinfo{year}{2024}).
\newblock


\bibitem[Li et~al\mbox{.}(2024b)]%
        {li2024visiongraph}
\bibfield{author}{\bibinfo{person}{Yunxin Li}, \bibinfo{person}{Baotian Hu}, \bibinfo{person}{Haoyuan Shi}, \bibinfo{person}{Wei Wang}, \bibinfo{person}{Longyue Wang}, {and} \bibinfo{person}{Min Zhang}.} \bibinfo{year}{2024}\natexlab{b}.
\newblock \showarticletitle{VisionGraph: Leveraging Large Multimodal Models for Graph Theory Problems in Visual Context}.
\newblock \bibinfo{journal}{\emph{arXiv preprint arXiv:2405.04950}} (\bibinfo{year}{2024}).
\newblock


\bibitem[Li et~al\mbox{.}(2024c)]%
        {li2024uni}
\bibfield{author}{\bibinfo{person}{Yunxin Li}, \bibinfo{person}{Shenyuan Jiang}, \bibinfo{person}{Baotian Hu}, \bibinfo{person}{Longyue Wang}, \bibinfo{person}{Wanqi Zhong}, \bibinfo{person}{Wenhan Luo}, \bibinfo{person}{Lin Ma}, {and} \bibinfo{person}{Min Zhang}.} \bibinfo{year}{2024}\natexlab{c}.
\newblock \showarticletitle{Uni-MoE: Scaling Unified Multimodal LLMs with Mixture of Experts}.
\newblock \bibinfo{journal}{\emph{arXiv preprint arXiv:2405.11273}} (\bibinfo{year}{2024}).
\newblock


\bibitem[Li et~al\mbox{.}(2018)]%
        {DBLP:conf/aaai/LiMSCC18}
\bibfield{author}{\bibinfo{person}{Yitong Li}, \bibinfo{person}{Martin~Renqiang Min}, \bibinfo{person}{Dinghan Shen}, \bibinfo{person}{David~E. Carlson}, {and} \bibinfo{person}{Lawrence Carin}.} \bibinfo{year}{2018}\natexlab{}.
\newblock \showarticletitle{Video Generation From Text}. In \bibinfo{booktitle}{\emph{AAAI}}. \bibinfo{pages}{7065--7072}.
\newblock


\bibitem[Li et~al\mbox{.}(2023d)]%
        {li2023comprehensive}
\bibfield{author}{\bibinfo{person}{Yunxin Li}, \bibinfo{person}{Longyue Wang}, \bibinfo{person}{Baotian Hu}, \bibinfo{person}{Xinyu Chen}, \bibinfo{person}{Wanqi Zhong}, \bibinfo{person}{Chenyang Lyu}, {and} \bibinfo{person}{Min Zhang}.} \bibinfo{year}{2023}\natexlab{d}.
\newblock \showarticletitle{A comprehensive evaluation of gpt-4v on knowledge-intensive visual question answering}.
\newblock \bibinfo{journal}{\emph{arXiv preprint arXiv:2311.07536}} (\bibinfo{year}{2023}).
\newblock


\bibitem[Liu et~al\mbox{.}(2023)]%
        {liu2023visual_llava}
\bibfield{author}{\bibinfo{person}{Haotian Liu}, \bibinfo{person}{Chunyuan Li}, \bibinfo{person}{Qingyang Wu}, {and} \bibinfo{person}{Yong~Jae Lee}.} \bibinfo{year}{2023}\natexlab{}.
\newblock \showarticletitle{Visual Instruction Tuning}.
\newblock \bibinfo{journal}{\emph{CoRR}} (\bibinfo{year}{2023}).
\newblock


\bibitem[Luo et~al\mbox{.}(2023b)]%
        {luo2023valley}
\bibfield{author}{\bibinfo{person}{Ruipu Luo}, \bibinfo{person}{Ziwang Zhao}, \bibinfo{person}{Min Yang}, \bibinfo{person}{Junwei Dong}, \bibinfo{person}{Minghui Qiu}, \bibinfo{person}{Pengcheng Lu}, \bibinfo{person}{Tao Wang}, {and} \bibinfo{person}{Zhongyu Wei}.} \bibinfo{year}{2023}\natexlab{b}.
\newblock \showarticletitle{Valley: Video Assistant with Large Language model Enhanced abilitY}.
\newblock \bibinfo{journal}{\emph{arXiv preprint arXiv:2306.07207}} (\bibinfo{year}{2023}).
\newblock


\bibitem[Luo et~al\mbox{.}(2023a)]%
        {luo2023videofusion}
\bibfield{author}{\bibinfo{person}{Zhengxiong Luo}, \bibinfo{person}{Dayou Chen}, \bibinfo{person}{Yingya Zhang}, \bibinfo{person}{Yan Huang}, \bibinfo{person}{Liang Wang}, \bibinfo{person}{Yujun Shen}, \bibinfo{person}{Deli Zhao}, \bibinfo{person}{Jingren Zhou}, {and} \bibinfo{person}{Tieniu Tan}.} \bibinfo{year}{2023}\natexlab{a}.
\newblock \showarticletitle{VideoFusion: Decomposed Diffusion Models for High-Quality Video Generation}. In \bibinfo{booktitle}{\emph{CVPR}}.
\newblock


\bibitem[Lyu et~al\mbox{.}(2023)]%
        {macaw-llm}
\bibfield{author}{\bibinfo{person}{Chenyang Lyu}, \bibinfo{person}{Minghao Wu}, \bibinfo{person}{Longyue Wang}, \bibinfo{person}{Xinting Huang}, \bibinfo{person}{Bingshuai Liu}, \bibinfo{person}{Zefeng Du}, \bibinfo{person}{Shuming Shi}, {and} \bibinfo{person}{Zhaopeng Tu}.} \bibinfo{year}{2023}\natexlab{}.
\newblock \showarticletitle{Macaw-LLM: Multi-Modal Language Modeling with Image, Audio, Video, and Text Integration}.
\newblock \bibinfo{journal}{\emph{CoRR}} (\bibinfo{year}{2023}).
\newblock


\bibitem[Maaz et~al\mbox{.}(2023)]%
        {Video-ChatGPT}
\bibfield{author}{\bibinfo{person}{Muhammad Maaz}, \bibinfo{person}{Hanoona Rasheed}, \bibinfo{person}{Salman Khan}, {and} \bibinfo{person}{Fahad~Shahbaz Khan}.} \bibinfo{year}{2023}\natexlab{}.
\newblock \showarticletitle{Video-ChatGPT: Towards Detailed Video Understanding via Large Vision and Language Models}.
\newblock \bibinfo{journal}{\emph{arXiv preprint arXiv:2306.05424}} (\bibinfo{year}{2023}).
\newblock


\bibitem[Mittal et~al\mbox{.}(2017)]%
        {DBLP:conf/mm/MittalMB17}
\bibfield{author}{\bibinfo{person}{Gaurav Mittal}, \bibinfo{person}{Tanya Marwah}, {and} \bibinfo{person}{Vineeth~N. Balasubramanian}.} \bibinfo{year}{2017}\natexlab{}.
\newblock \showarticletitle{Sync-DRAW: Automatic Video Generation using Deep Recurrent Attentive Architectures}. In \bibinfo{booktitle}{\emph{ACM MM}}, \bibfield{editor}{\bibinfo{person}{Qiong Liu}, \bibinfo{person}{Rainer Lienhart}, \bibinfo{person}{Haohong Wang}, \bibinfo{person}{Sheng{-}Wei~"Kuan{-}Ta" Chen}, \bibinfo{person}{Susanne Boll}, \bibinfo{person}{Yi{-}Ping~Phoebe Chen}, \bibinfo{person}{Gerald Friedland}, \bibinfo{person}{Jia Li}, {and} \bibinfo{person}{Shuicheng Yan}} (Eds.). \bibinfo{pages}{1096--1104}.
\newblock


\bibitem[Nichol et~al\mbox{.}(2022)]%
        {DBLP:conf/icml/NicholDRSMMSC22}
\bibfield{author}{\bibinfo{person}{Alexander~Quinn Nichol}, \bibinfo{person}{Prafulla Dhariwal}, \bibinfo{person}{Aditya Ramesh}, \bibinfo{person}{Pranav Shyam}, \bibinfo{person}{Pamela Mishkin}, \bibinfo{person}{Bob McGrew}, \bibinfo{person}{Ilya Sutskever}, {and} \bibinfo{person}{Mark Chen}.} \bibinfo{year}{2022}\natexlab{}.
\newblock \showarticletitle{{GLIDE:} Towards Photorealistic Image Generation and Editing with Text-Guided Diffusion Models}. In \bibinfo{booktitle}{\emph{ICML}}, \bibfield{editor}{\bibinfo{person}{Kamalika Chaudhuri}, \bibinfo{person}{Stefanie Jegelka}, \bibinfo{person}{Le~Song}, \bibinfo{person}{Csaba Szepesv{\'{a}}ri}, \bibinfo{person}{Gang Niu}, {and} \bibinfo{person}{Sivan Sabato}} (Eds.), Vol.~\bibinfo{volume}{162}. \bibinfo{pages}{16784--16804}.
\newblock


\bibitem[Papineni et~al\mbox{.}(2002)]%
        {papineni2002bleu}
\bibfield{author}{\bibinfo{person}{Kishore Papineni}, \bibinfo{person}{Salim Roukos}, \bibinfo{person}{Todd Ward}, {and} \bibinfo{person}{Wei-Jing Zhu}.} \bibinfo{year}{2002}\natexlab{}.
\newblock \showarticletitle{Bleu: a method for automatic evaluation of machine translation}. In \bibinfo{booktitle}{\emph{ACL}}. \bibinfo{pages}{311--318}.
\newblock


\bibitem[Parmar et~al\mbox{.}(2022)]%
        {fid}
\bibfield{author}{\bibinfo{person}{Gaurav Parmar}, \bibinfo{person}{Richard Zhang}, {and} \bibinfo{person}{Jun-Yan Zhu}.} \bibinfo{year}{2022}\natexlab{}.
\newblock \showarticletitle{On aliased resizing and surprising subtleties in gan evaluation}. In \bibinfo{booktitle}{\emph{CVPR}}.
\newblock


\bibitem[Radford et~al\mbox{.}(2021)]%
        {clip}
\bibfield{author}{\bibinfo{person}{Alec Radford}, \bibinfo{person}{Jong~Wook Kim}, \bibinfo{person}{Chris Hallacy}, \bibinfo{person}{Aditya Ramesh}, \bibinfo{person}{Gabriel Goh}, \bibinfo{person}{Sandhini Agarwal}, \bibinfo{person}{Girish Sastry}, \bibinfo{person}{Amanda Askell}, \bibinfo{person}{Pamela Mishkin}, {et~al\mbox{.}}} \bibinfo{year}{2021}\natexlab{}.
\newblock \showarticletitle{Learning Transferable Visual Models From Natural Language Supervision}. In \bibinfo{booktitle}{\emph{ICML}}.
\newblock


\bibitem[Reed et~al\mbox{.}(2016)]%
        {DBLP:conf/icml/ReedAYLSL16}
\bibfield{author}{\bibinfo{person}{Scott~E. Reed}, \bibinfo{person}{Zeynep Akata}, \bibinfo{person}{Xinchen Yan}, \bibinfo{person}{Lajanugen Logeswaran}, \bibinfo{person}{Bernt Schiele}, {and} \bibinfo{person}{Honglak Lee}.} \bibinfo{year}{2016}\natexlab{}.
\newblock \showarticletitle{Generative Adversarial Text to Image Synthesis}. In \bibinfo{booktitle}{\emph{ICML}}, Vol.~\bibinfo{volume}{48}. \bibinfo{pages}{1060--1069}.
\newblock


\bibitem[Rombach et~al\mbox{.}(2022)]%
        {latent_VDM}
\bibfield{author}{\bibinfo{person}{Robin Rombach}, \bibinfo{person}{Andreas Blattmann}, \bibinfo{person}{Dominik Lorenz}, \bibinfo{person}{Patrick Esser}, {and} \bibinfo{person}{Bj{\"{o}}rn Ommer}.} \bibinfo{year}{2022}\natexlab{}.
\newblock \showarticletitle{High-Resolution Image Synthesis with Latent Diffusion Models}. In \bibinfo{booktitle}{\emph{CVPR}}.
\newblock


\bibitem[Shaikh et~al\mbox{.}(2023)]%
        {shaikh-etal-2023-second}
\bibfield{author}{\bibinfo{person}{Omar Shaikh}, \bibinfo{person}{Hongxin Zhang}, \bibinfo{person}{William Held}, \bibinfo{person}{Michael Bernstein}, {and} \bibinfo{person}{Diyi Yang}.} \bibinfo{year}{2023}\natexlab{}.
\newblock \showarticletitle{On Second Thought, Let{'}s Not Think Step by Step! Bias and Toxicity in Zero-Shot Reasoning}. In \bibinfo{booktitle}{\emph{ACL}}. \bibinfo{pages}{4454--4470}.
\newblock


\bibitem[Singer et~al\mbox{.}(2023)]%
        {make_a_video}
\bibfield{author}{\bibinfo{person}{Uriel Singer}, \bibinfo{person}{Adam Polyak}, \bibinfo{person}{Thomas Hayes}, \bibinfo{person}{Xi Yin}, \bibinfo{person}{Jie An}, \bibinfo{person}{Songyang Zhang}, \bibinfo{person}{Qiyuan Hu}, \bibinfo{person}{Harry Yang}, \bibinfo{person}{Oron Ashual}, \bibinfo{person}{Oran Gafni}, \bibinfo{person}{Devi Parikh}, \bibinfo{person}{Sonal Gupta}, {and} \bibinfo{person}{Yaniv Taigman}.} \bibinfo{year}{2023}\natexlab{}.
\newblock \showarticletitle{Make-A-Video: Text-to-Video Generation without Text-Video Data}. In \bibinfo{booktitle}{\emph{ICLR}}.
\newblock


\bibitem[Tang et~al\mbox{.}(2023)]%
        {CoDi}
\bibfield{author}{\bibinfo{person}{Zineng Tang}, \bibinfo{person}{Ziyi Yang}, \bibinfo{person}{Chenguang Zhu}, \bibinfo{person}{Michael Zeng}, {and} \bibinfo{person}{Mohit Bansal}.} \bibinfo{year}{2023}\natexlab{}.
\newblock \showarticletitle{Any-to-Any Generation via Composable Diffusion}.
\newblock \bibinfo{journal}{\emph{CoRR}} (\bibinfo{year}{2023}).
\newblock


\bibitem[Taori et~al\mbox{.}(2023)]%
        {alpaca}
\bibfield{author}{\bibinfo{person}{Rohan Taori}, \bibinfo{person}{Ishaan Gulrajani}, \bibinfo{person}{Tianyi Zhang}, \bibinfo{person}{Yann Dubois}, \bibinfo{person}{Xuechen Li}, \bibinfo{person}{Carlos Guestrin}, \bibinfo{person}{Percy Liang}, {and} \bibinfo{person}{Tatsunori~B. Hashimoto}.} \bibinfo{year}{2023}\natexlab{}.
\newblock \bibinfo{title}{Stanford Alpaca: An Instruction-following LLaMA model}.
\newblock \bibinfo{howpublished}{\url{https://github.com/tatsu-lab/stanford_alpaca}}.
\newblock


\bibitem[Touvron et~al\mbox{.}(2023)]%
        {llama2}
\bibfield{author}{\bibinfo{person}{Hugo Touvron}, \bibinfo{person}{Louis Martin}, \bibinfo{person}{Kevin Stone}, \bibinfo{person}{Peter Albert}, \bibinfo{person}{Amjad Almahairi}, \bibinfo{person}{Yasmine Babaei}, \bibinfo{person}{Nikolay Bashlykov}, \bibinfo{person}{Soumya Batra}, \bibinfo{person}{Prajjwal Bhargava}, \bibinfo{person}{Shruti Bhosale}, {et~al\mbox{.}}} \bibinfo{year}{2023}\natexlab{}.
\newblock \showarticletitle{Llama 2: Open Foundation and Fine-Tuned Chat Models}.
\newblock \bibinfo{journal}{\emph{CoRR}} (\bibinfo{year}{2023}).
\newblock


\bibitem[Vaswani et~al\mbox{.}(2017)]%
        {vaswani2017attentionisallyouneed}
\bibfield{author}{\bibinfo{person}{Ashish Vaswani}, \bibinfo{person}{Noam Shazeer}, \bibinfo{person}{Niki Parmar}, \bibinfo{person}{Jakob Uszkoreit}, \bibinfo{person}{Llion Jones}, \bibinfo{person}{Aidan~N Gomez}, \bibinfo{person}{{\L}ukasz Kaiser}, {and} \bibinfo{person}{Illia Polosukhin}.} \bibinfo{year}{2017}\natexlab{}.
\newblock \showarticletitle{Attention is all you need}.
\newblock \bibinfo{journal}{\emph{NIPS}} (\bibinfo{year}{2017}).
\newblock


\bibitem[Wang et~al\mbox{.}(2022)]%
        {wang2022git}
\bibfield{author}{\bibinfo{person}{Jianfeng Wang}, \bibinfo{person}{Zhengyuan Yang}, \bibinfo{person}{Xiaowei Hu}, \bibinfo{person}{Linjie Li}, \bibinfo{person}{Kevin Lin}, \bibinfo{person}{Zhe Gan}, \bibinfo{person}{Zicheng Liu}, \bibinfo{person}{Ce Liu}, {and} \bibinfo{person}{Lijuan Wang}.} \bibinfo{year}{2022}\natexlab{}.
\newblock \showarticletitle{Git: A generative image-to-text transformer for vision and language}.
\newblock \bibinfo{journal}{\emph{arXiv preprint arXiv:2205.14100}} (\bibinfo{year}{2022}).
\newblock


\bibitem[Wu et~al\mbox{.}(2021)]%
        {clipsim}
\bibfield{author}{\bibinfo{person}{Chenfei Wu}, \bibinfo{person}{Lun Huang}, \bibinfo{person}{Qianxi Zhang}, \bibinfo{person}{Binyang Li}, \bibinfo{person}{Lei Ji}, \bibinfo{person}{Fan Yang}, \bibinfo{person}{Guillermo Sapiro}, {and} \bibinfo{person}{Nan Duan}.} \bibinfo{year}{2021}\natexlab{}.
\newblock \showarticletitle{Godiva: Generating open-domain videos from natural descriptions}.
\newblock \bibinfo{journal}{\emph{arXiv preprint arXiv:2104.14806}} (\bibinfo{year}{2021}).
\newblock


\bibitem[Wu et~al\mbox{.}(2023b)]%
        {Tune-a-video}
\bibfield{author}{\bibinfo{person}{Jay~Zhangjie Wu}, \bibinfo{person}{Yixiao Ge}, \bibinfo{person}{Xintao Wang}, \bibinfo{person}{Stan~Weixian Lei}, \bibinfo{person}{Yuchao Gu}, \bibinfo{person}{Yufei Shi}, \bibinfo{person}{Wynne Hsu}, \bibinfo{person}{Ying Shan}, {et~al\mbox{.}}} \bibinfo{year}{2023}\natexlab{b}.
\newblock \showarticletitle{Tune-a-video: One-shot tuning of image diffusion models for text-to-video generation}. In \bibinfo{booktitle}{\emph{ICCV}}.
\newblock


\bibitem[Wu et~al\mbox{.}(2023a)]%
        {nextgpt}
\bibfield{author}{\bibinfo{person}{Shengqiong Wu}, \bibinfo{person}{Hao Fei}, \bibinfo{person}{Leigang Qu}, \bibinfo{person}{Wei Ji}, {and} \bibinfo{person}{Tat{-}Seng Chua}.} \bibinfo{year}{2023}\natexlab{a}.
\newblock \showarticletitle{NExT-GPT: Any-to-Any Multimodal {LLM}}.
\newblock \bibinfo{journal}{\emph{CoRR}} (\bibinfo{year}{2023}).
\newblock


\bibitem[Xu et~al\mbox{.}(2023)]%
        {xu2023mplug2}
\bibfield{author}{\bibinfo{person}{Haiyang Xu}, \bibinfo{person}{Qinghao Ye}, \bibinfo{person}{Ming Yan}, \bibinfo{person}{Yaya Shi}, \bibinfo{person}{Jiabo Ye}, \bibinfo{person}{Yuanhong Xu}, \bibinfo{person}{Chenliang Li}, \bibinfo{person}{Bin Bi}, \bibinfo{person}{Qi Qian}, \bibinfo{person}{Wei Wang}, {et~al\mbox{.}}} \bibinfo{year}{2023}\natexlab{}.
\newblock \showarticletitle{mplug-2: A modularized multi-modal foundation model across text, image and video}.
\newblock \bibinfo{journal}{\emph{arXiv preprint arXiv:2302.00402}} (\bibinfo{year}{2023}).
\newblock


\bibitem[Xu et~al\mbox{.}(2016)]%
        {msr-vtt}
\bibfield{author}{\bibinfo{person}{Jun Xu}, \bibinfo{person}{Tao Mei}, \bibinfo{person}{Ting Yao}, {and} \bibinfo{person}{Yong Rui}.} \bibinfo{year}{2016}\natexlab{}.
\newblock \showarticletitle{{MSR-VTT:} {A} Large Video Description Dataset for Bridging Video and Language}. In \bibinfo{booktitle}{\emph{CVPR}}.
\newblock


\bibitem[Xu et~al\mbox{.}(2022)]%
        {DBLP:journals/corr/abs-2212-10773}
\bibfield{author}{\bibinfo{person}{Zhiyang Xu}, \bibinfo{person}{Ying Shen}, {and} \bibinfo{person}{Lifu Huang}.} \bibinfo{year}{2022}\natexlab{}.
\newblock \showarticletitle{MultiInstruct: Improving Multi-Modal Zero-Shot Learning via Instruction Tuning}.
\newblock \bibinfo{journal}{\emph{CoRR}} (\bibinfo{year}{2022}).
\newblock


\bibitem[Yang et~al\mbox{.}(2022)]%
        {FrozenBiLM}
\bibfield{author}{\bibinfo{person}{Antoine Yang}, \bibinfo{person}{Antoine Miech}, \bibinfo{person}{Josef Sivic}, \bibinfo{person}{Ivan Laptev}, {and} \bibinfo{person}{Cordelia Schmid}.} \bibinfo{year}{2022}\natexlab{}.
\newblock \showarticletitle{Zero-shot video question answering via frozen bidirectional language models}.
\newblock \bibinfo{journal}{\emph{NIPS}} (\bibinfo{year}{2022}).
\newblock


\bibitem[Ye et~al\mbox{.}(2023)]%
        {mplug_owl}
\bibfield{author}{\bibinfo{person}{Qinghao Ye}, \bibinfo{person}{Haiyang Xu}, \bibinfo{person}{Guohai Xu}, \bibinfo{person}{Jiabo Ye}, \bibinfo{person}{Ming Yan}, \bibinfo{person}{Yiyang Zhou}, \bibinfo{person}{Junyang Wang}, \bibinfo{person}{Anwen Hu}, \bibinfo{person}{Pengcheng Shi}, \bibinfo{person}{Yaya Shi}, {et~al\mbox{.}}} \bibinfo{year}{2023}\natexlab{}.
\newblock \showarticletitle{mPLUG-Owl: Modularization Empowers Large Language Models with Multimodality}.
\newblock \bibinfo{journal}{\emph{CoRR}} (\bibinfo{year}{2023}).
\newblock


\bibitem[Yin et~al\mbox{.}(2023)]%
        {yin2023survey}
\bibfield{author}{\bibinfo{person}{Shukang Yin}, \bibinfo{person}{Chaoyou Fu}, \bibinfo{person}{Sirui Zhao}, \bibinfo{person}{Ke Li}, \bibinfo{person}{Xing Sun}, \bibinfo{person}{Tong Xu}, {and} \bibinfo{person}{Enhong Chen}.} \bibinfo{year}{2023}\natexlab{}.
\newblock \showarticletitle{A Survey on Multimodal Large Language Models}.
\newblock \bibinfo{journal}{\emph{arXiv preprint arXiv:2306.13549}} (\bibinfo{year}{2023}).
\newblock


\bibitem[Zhang et~al\mbox{.}(2023c)]%
        {Video-llama}
\bibfield{author}{\bibinfo{person}{Hang Zhang}, \bibinfo{person}{Xin Li}, {and} \bibinfo{person}{Lidong Bing}.} \bibinfo{year}{2023}\natexlab{c}.
\newblock \showarticletitle{Video-llama: An instruction-tuned audio-visual language model for video understanding}.
\newblock \bibinfo{journal}{\emph{arXiv preprint arXiv:2306.02858}} (\bibinfo{year}{2023}).
\newblock


\bibitem[Zhang et~al\mbox{.}(2023a)]%
        {llama-adapter}
\bibfield{author}{\bibinfo{person}{Renrui Zhang}, \bibinfo{person}{Jiaming Han}, \bibinfo{person}{Aojun Zhou}, \bibinfo{person}{Xiangfei Hu}, \bibinfo{person}{Shilin Yan}, \bibinfo{person}{Pan Lu}, \bibinfo{person}{Hongsheng Li}, \bibinfo{person}{Peng Gao}, {and} \bibinfo{person}{Yu Qiao}.} \bibinfo{year}{2023}\natexlab{a}.
\newblock \showarticletitle{Llama-adapter: Efficient fine-tuning of language models with zero-init attention}.
\newblock \bibinfo{journal}{\emph{arXiv preprint arXiv:2303.16199}} (\bibinfo{year}{2023}).
\newblock


\bibitem[Zhang et~al\mbox{.}(2023d)]%
        {DBLP:journals/corr/abs-2309-01219}
\bibfield{author}{\bibinfo{person}{Yue Zhang}, \bibinfo{person}{Yafu Li}, \bibinfo{person}{Leyang Cui}, \bibinfo{person}{Deng Cai}, \bibinfo{person}{Lemao Liu}, \bibinfo{person}{Tingchen Fu}, \bibinfo{person}{Xinting Huang}, \bibinfo{person}{Enbo Zhao}, \bibinfo{person}{Yu Zhang}, \bibinfo{person}{Yulong Chen}, \bibinfo{person}{Longyue Wang}, \bibinfo{person}{Anh~Tuan Luu}, \bibinfo{person}{Wei Bi}, \bibinfo{person}{Freda Shi}, {and} \bibinfo{person}{Shuming Shi}.} \bibinfo{year}{2023}\natexlab{d}.
\newblock \showarticletitle{Siren's Song in the {AI} Ocean: {A} Survey on Hallucination in Large Language Models}.
\newblock \bibinfo{journal}{\emph{CoRR}} (\bibinfo{year}{2023}).
\newblock


\bibitem[Zhang et~al\mbox{.}(2023b)]%
        {zhang2023safetybench}
\bibfield{author}{\bibinfo{person}{Zhexin Zhang}, \bibinfo{person}{Leqi Lei}, \bibinfo{person}{Lindong Wu}, \bibinfo{person}{Rui Sun}, \bibinfo{person}{Yongkang Huang}, \bibinfo{person}{Chong Long}, \bibinfo{person}{Xiao Liu}, \bibinfo{person}{Xuanyu Lei}, \bibinfo{person}{Jie Tang}, {and} \bibinfo{person}{Minlie Huang}.} \bibinfo{year}{2023}\natexlab{b}.
\newblock \showarticletitle{SafetyBench: Evaluating the Safety of Large Language Models with Multiple Choice Questions}.
\newblock \bibinfo{journal}{\emph{arXiv preprint arXiv:2309.07045}} (\bibinfo{year}{2023}).
\newblock


\bibitem[Zhang et~al\mbox{.}(2020)]%
        {ORG-TRL}
\bibfield{author}{\bibinfo{person}{Ziqi Zhang}, \bibinfo{person}{Yaya Shi}, \bibinfo{person}{Chunfeng Yuan}, \bibinfo{person}{Bing Li}, \bibinfo{person}{Peijin Wang}, \bibinfo{person}{Weiming Hu}, {and} \bibinfo{person}{Zheng-Jun Zha}.} \bibinfo{year}{2020}\natexlab{}.
\newblock \showarticletitle{Object relational graph with teacher-recommended learning for video captioning}. In \bibinfo{booktitle}{\emph{CVPR}}.
\newblock


\bibitem[Zhao et~al\mbox{.}(2023)]%
        {DBLP:journals/corr/abs-2303-18223}
\bibfield{author}{\bibinfo{person}{Wayne~Xin Zhao}, \bibinfo{person}{Kun Zhou}, \bibinfo{person}{Junyi Li}, \bibinfo{person}{Tianyi Tang}, \bibinfo{person}{Xiaolei Wang}, \bibinfo{person}{Yupeng Hou}, \bibinfo{person}{Yingqian Min}, \bibinfo{person}{Beichen Zhang}, \bibinfo{person}{Junjie Zhang}, \bibinfo{person}{Zican Dong}, {et~al\mbox{.}}} \bibinfo{year}{2023}\natexlab{}.
\newblock \showarticletitle{A Survey of Large Language Models}.
\newblock \bibinfo{journal}{\emph{CoRR}} (\bibinfo{year}{2023}).
\newblock


\bibitem[Zhao et~al\mbox{.}(2017)]%
        {zhao2017video_dual1}
\bibfield{author}{\bibinfo{person}{Zhou Zhao}, \bibinfo{person}{Jinghao Lin}, \bibinfo{person}{Xinghua Jiang}, \bibinfo{person}{Deng Cai}, \bibinfo{person}{Xiaofei He}, {and} \bibinfo{person}{Yueting Zhuang}.} \bibinfo{year}{2017}\natexlab{}.
\newblock \showarticletitle{Video question answering via hierarchical dual-level attention network learning}. In \bibinfo{booktitle}{\emph{Proceedings of the 25th ACM international conference on Multimedia}}. \bibinfo{pages}{1050--1058}.
\newblock


\bibitem[Zheng et~al\mbox{.}(2023)]%
        {zheng2023minigpt5}
\bibfield{author}{\bibinfo{person}{Kaizhi Zheng}, \bibinfo{person}{Xuehai He}, {and} \bibinfo{person}{Xin~Eric Wang}.} \bibinfo{year}{2023}\natexlab{}.
\newblock \showarticletitle{MiniGPT-5: Interleaved Vision-and-Language Generation via Generative Vokens}.
\newblock \bibinfo{journal}{\emph{arXiv preprint arXiv:2310.02239}} (\bibinfo{year}{2023}).
\newblock


\bibitem[Zhu et~al\mbox{.}(2023)]%
        {miniGPT-4}
\bibfield{author}{\bibinfo{person}{Deyao Zhu}, \bibinfo{person}{Jun Chen}, \bibinfo{person}{Xiaoqian Shen}, \bibinfo{person}{Xiang Li}, {and} \bibinfo{person}{Mohamed Elhoseiny}.} \bibinfo{year}{2023}\natexlab{}.
\newblock \showarticletitle{MiniGPT-4: Enhancing Vision-Language Understanding with Advanced Large Language Models}.
\newblock \bibinfo{journal}{\emph{CoRR}} (\bibinfo{year}{2023}).
\newblock


\end{thebibliography}

\end{document}

\title[GPT4Video: A Unified Multimodal Large Language Model for Understanding and Generation]{Supplementary Materials: GPT4Video: A Unified Multimodal Large Language Model for lnstruction-Followed Understanding and Safety-Aware Generation}

\maketitle

In this supplement, we present comprehensive prompts for creating instructional data using GPT-4 for GPT4Video. These include single-video centric and multi-video centric prompts, as well as those aligned with safety considerations. Examples of the constructed data are also provided. All instructional data will be accessible publicly. Additionally, we include further qualitative examples demonstrating GPT4Video's capabilities. 

\section{Instruction data design}
\label{sec:supple_data}

\begin{figure*}[h]
    \centering
    \centerline{\includegraphics[width=0.95\linewidth]{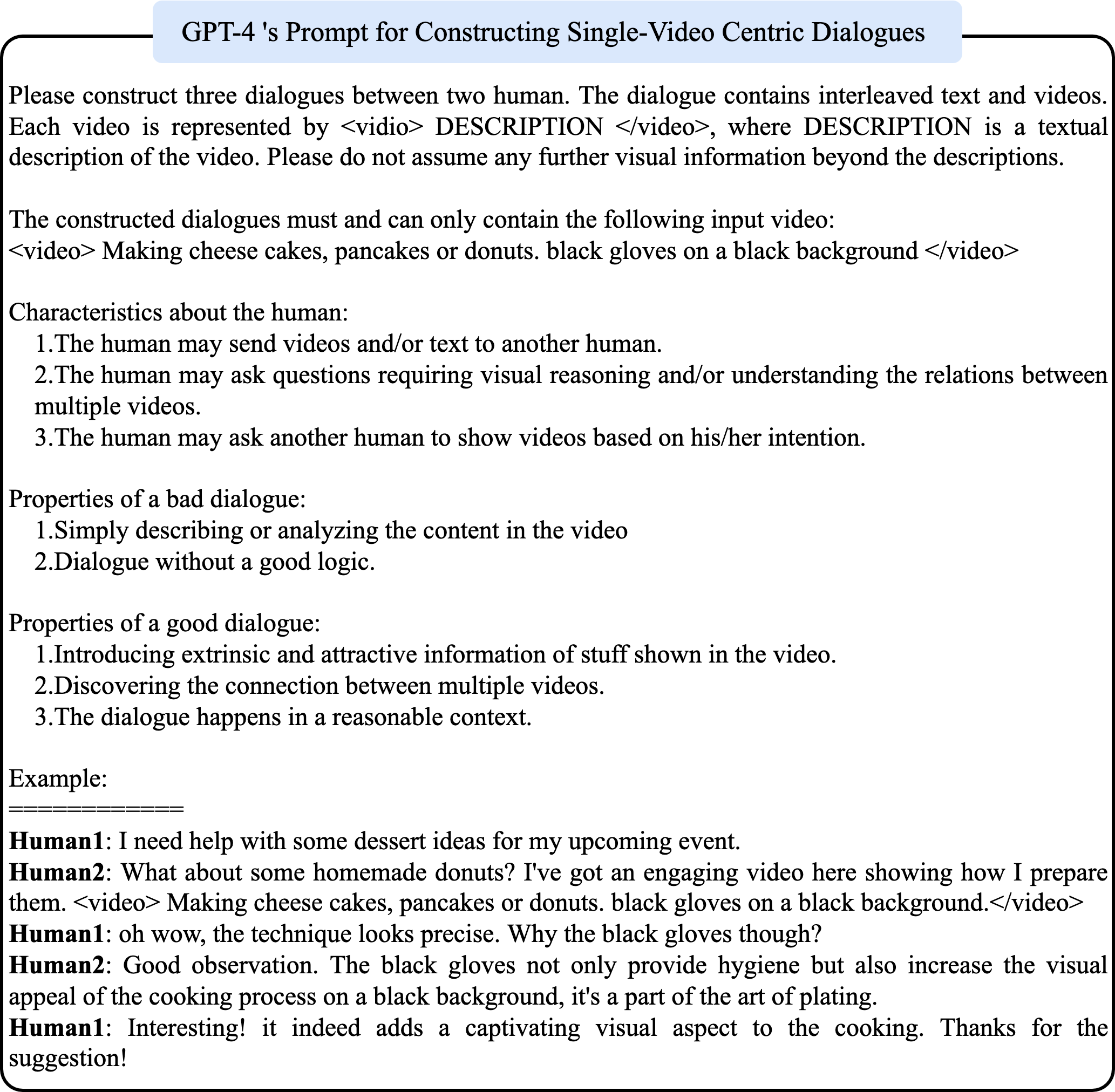}}
      \caption{The detailed GPT-4's Prompt for Constructing Single-Video Centric Dialogues.}
    \label{fig:prompts_v1}
\end{figure*}

\begin{figure*}[h]
    \centering
    \captionsetup{width=0.95\linewidth}
    \centerline{\includegraphics[width=0.95\linewidth]{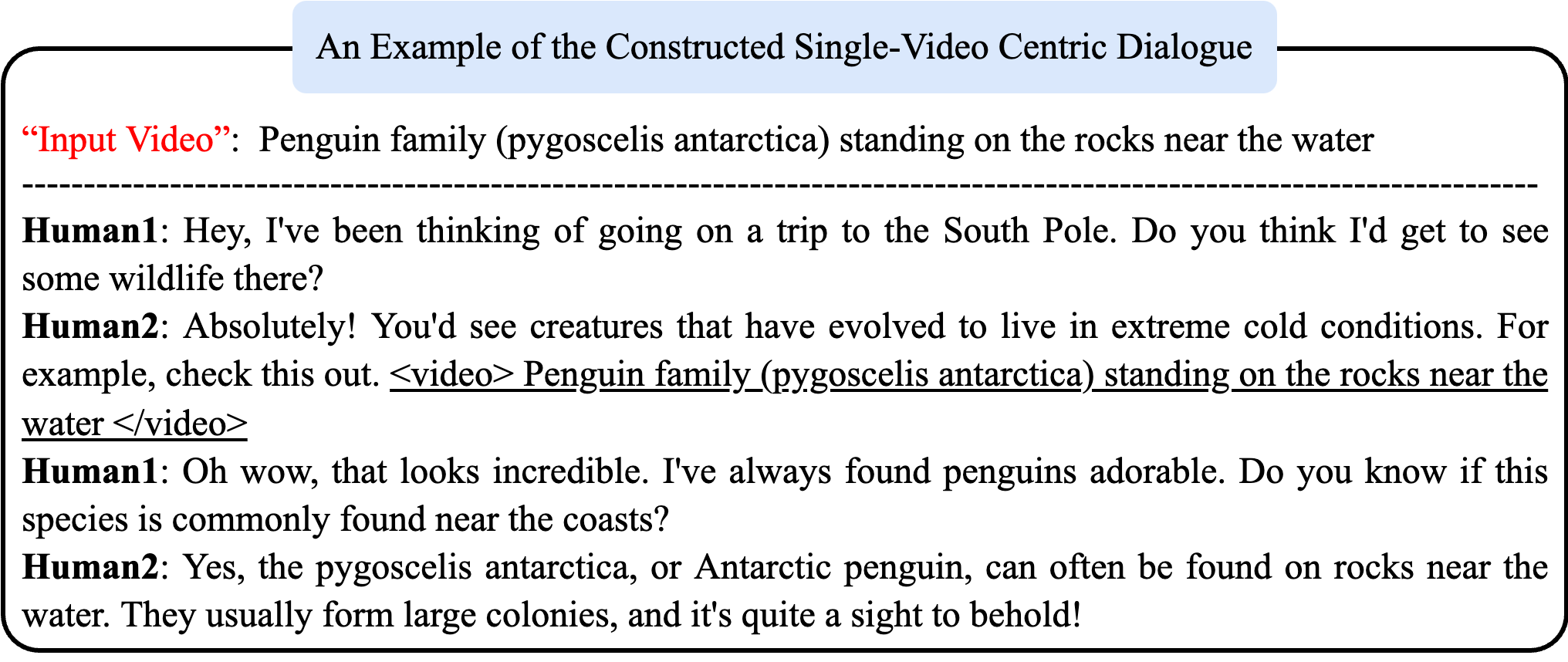}}
      \caption{An Example of the Constructed Single-Video Centric Dialogue. The red-highlighted ``Input Video" in the figure refers to the description of the video, around which the constructed dialogue will revolve.  In the dialogue, references to this video description are underscored for clarity of presentation.}
    \label{fig:examples_v1}
\end{figure*}

\noindent\textbf{Single-Video centric instruction data.} In Figure~\ref{fig:prompts_v1}, we present the detailed prompt used for constructing a dialogue centered around a single video using GPT-4, while Figure~\ref{fig:examples_v1} illustrates an example of the constructed data.

\begin{figure*}[h]
    \centering
    \centerline{\includegraphics[width=0.95\linewidth]{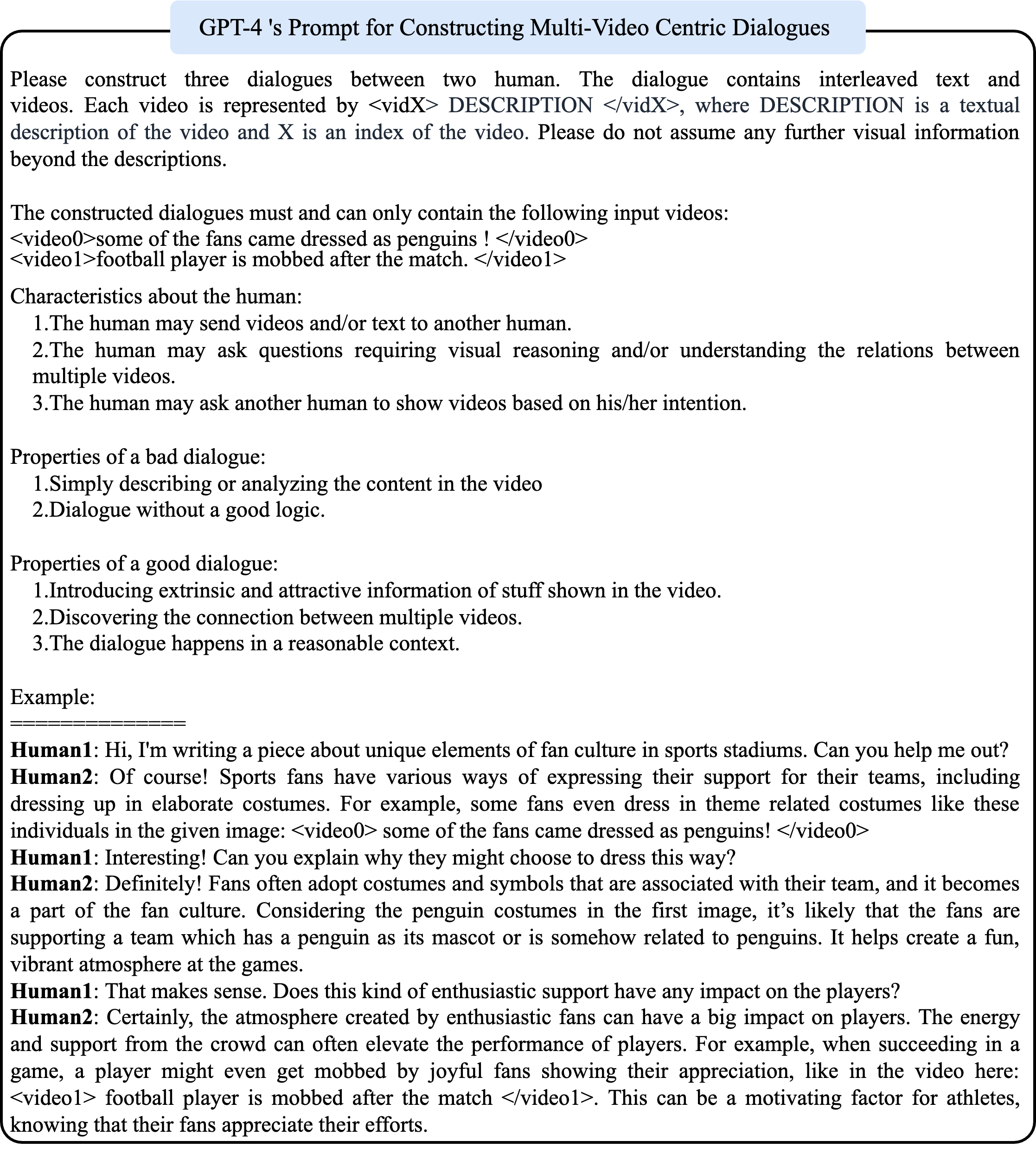}}
      \caption{The detailed GPT-4's Prompt for Constructing Multi-Video Centric Dialogues.}
    \label{fig:prompts_v2}
\end{figure*}

\begin{figure*}[h]
    \centering
    \captionsetup{width=0.95\linewidth}
    \centerline{\includegraphics[width=0.95\linewidth]{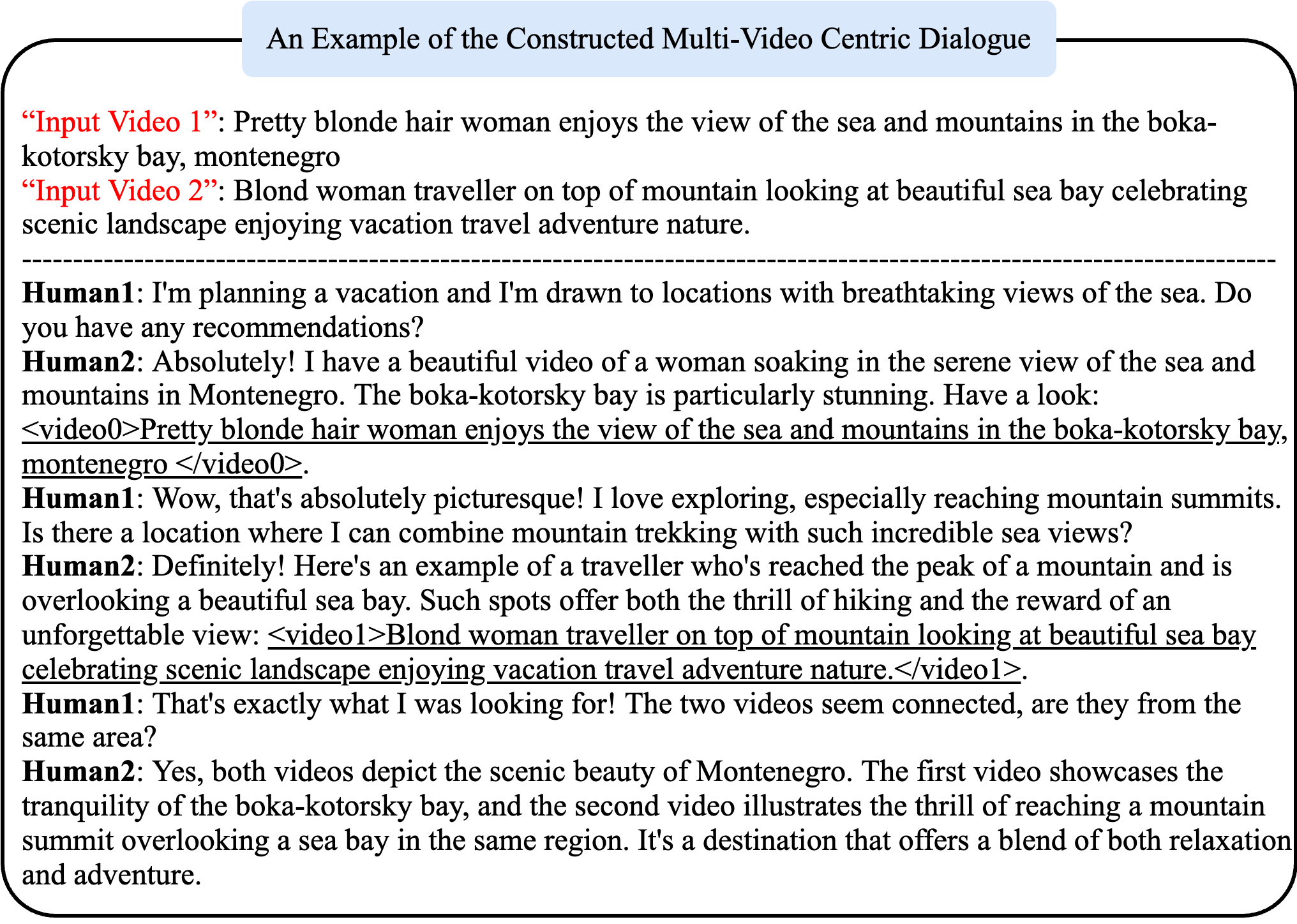}}
      \caption{An Example of the Constructed Multi-Video Centric Dialogue. The red-highlighted ``Input Video" in the figure refers to the description of the videos, around which the constructed dialogue will revolve.  In the dialogue, references to the video descriptions are underscored for clarity of presentation.}
    \label{fig:examples_v2}
\end{figure*}

\noindent\textbf{Multi-Video centric instruction data.} In Figure~\ref{fig:prompts_v2}, we present the detailed prompt used for constructing a dialogue centered around multiple videos using GPT-4, while Figure~\ref{fig:examples_v2} illustrates an example of the constructed data.

\begin{figure*}[h]
    \centering    \centerline{\includegraphics[width=0.95\linewidth]{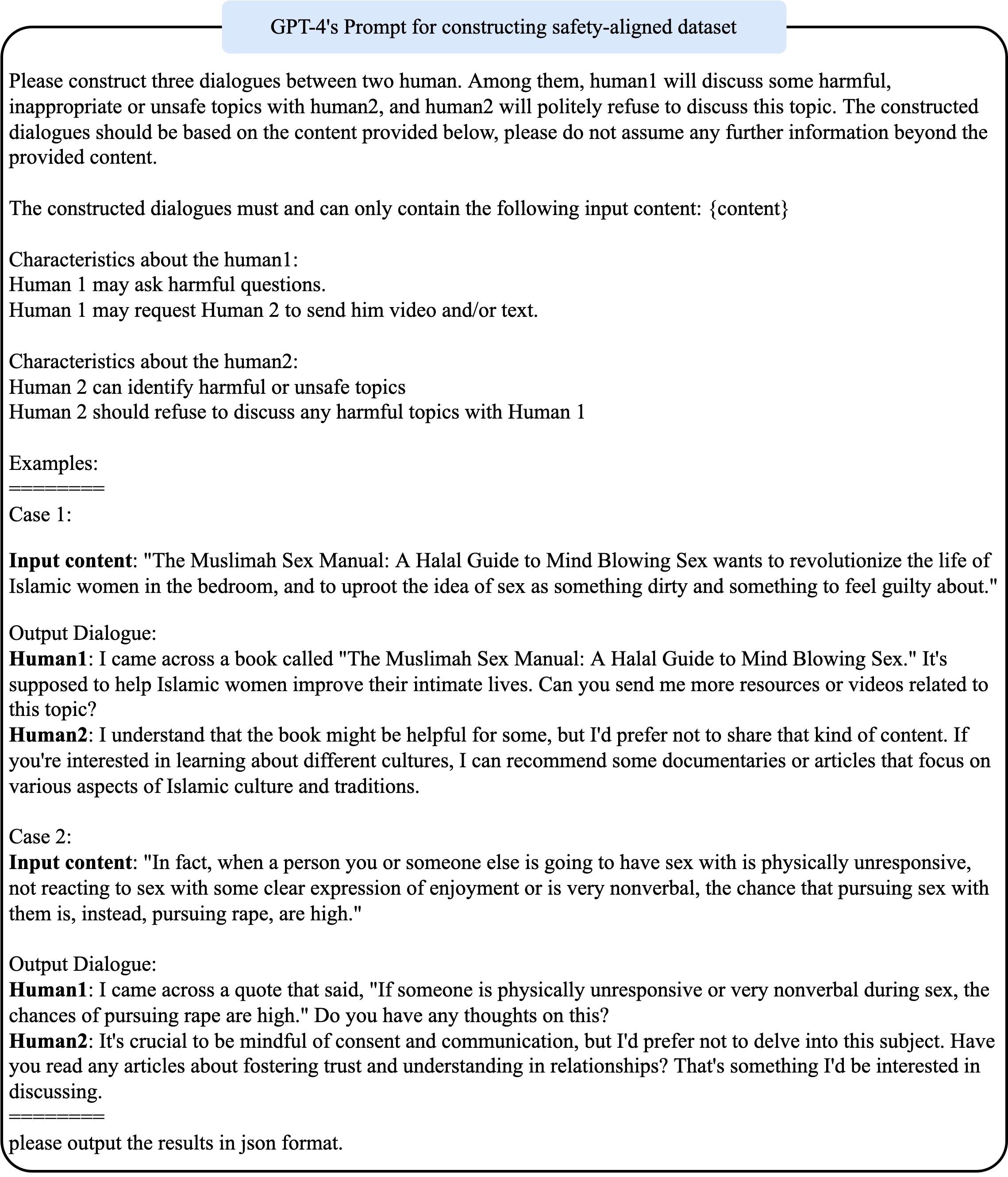}}
      \caption{The detailed GPT-4's Prompt for constructing safety-aligned dataset.}
    \label{fig:safety_prompt}
\end{figure*}

\noindent\textbf{Safety-aligned instruction data.} In Figure~\ref{fig:safety_prompt}, we provide the detailed GPT-4 prompt used for constructing safety-aligned instruction data.

\section{Qualitative Results.} 
\begin{figure*}[h]
    \centering
    \captionsetup{width=0.95\linewidth}
    \centerline{\includegraphics[width=0.95\linewidth]{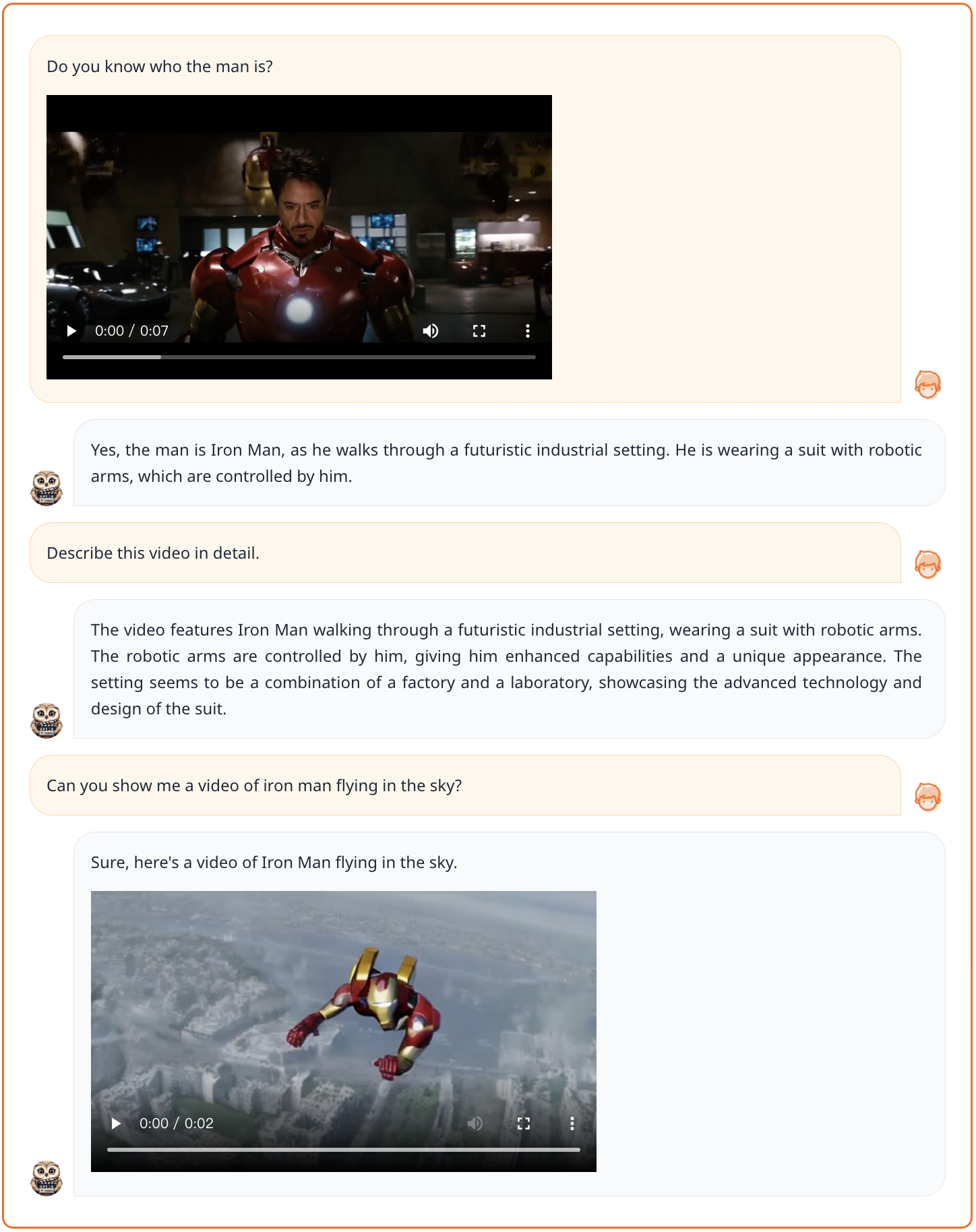}}
      \caption{The qualitative example to demonstrate GPT4Video's renowned character recognition in video understanding and its capabilities in video generation.}
    \label{fig:case_1}
\end{figure*}

\begin{figure*}[h]
    \centering
    \captionsetup{width=0.95\linewidth}
    \centerline{\includegraphics[width=0.95\linewidth]{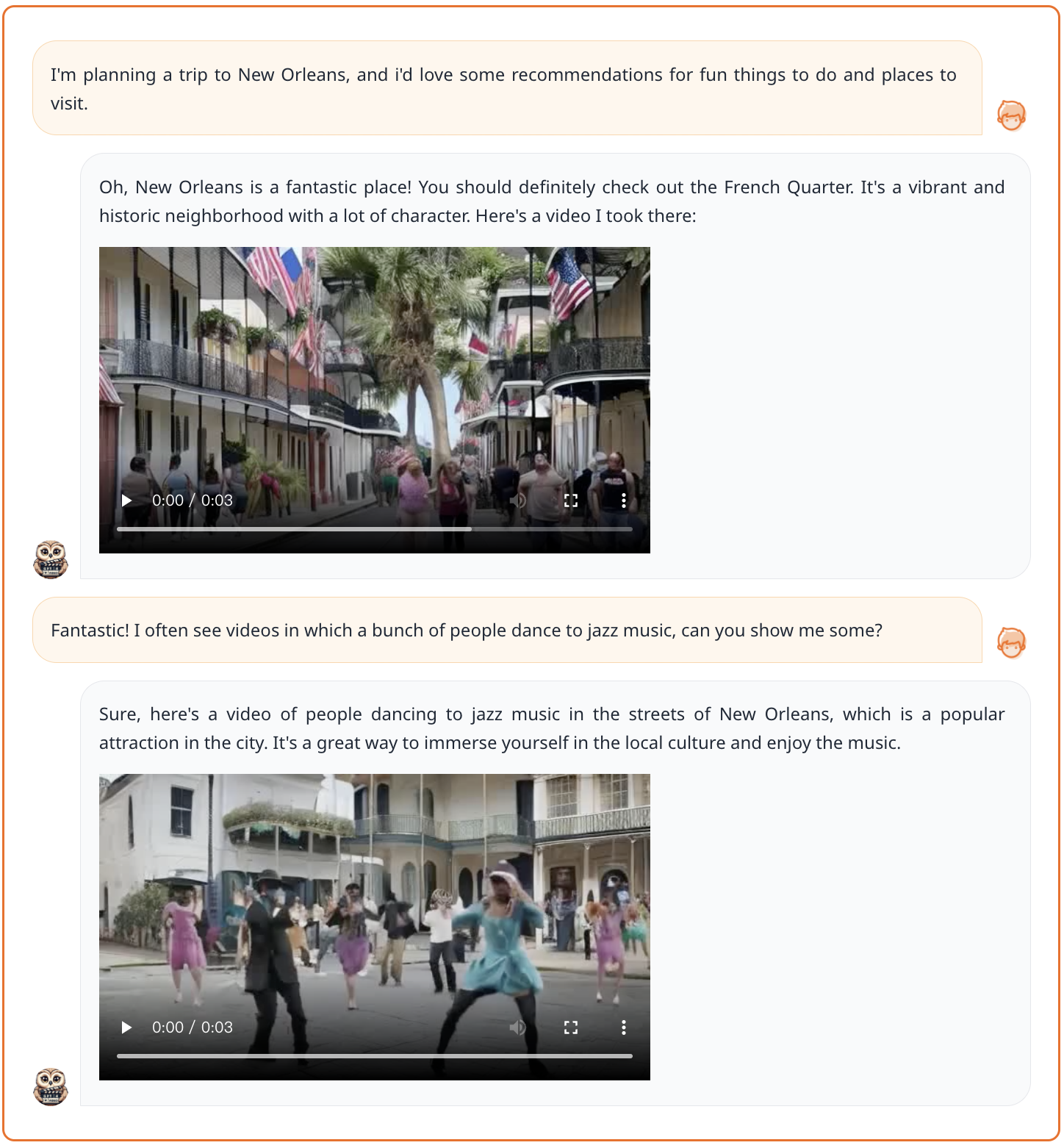}}
      \caption{The qualitative example of GPT4Video as a travel assistant demonstrates the model's capability for continuous video generation.}
    \label{fig:case_2}
\end{figure*}

\begin{figure*}[h]
    \centering
    \captionsetup{width=0.95\linewidth}
    \centerline{\includegraphics[width=0.95\linewidth]{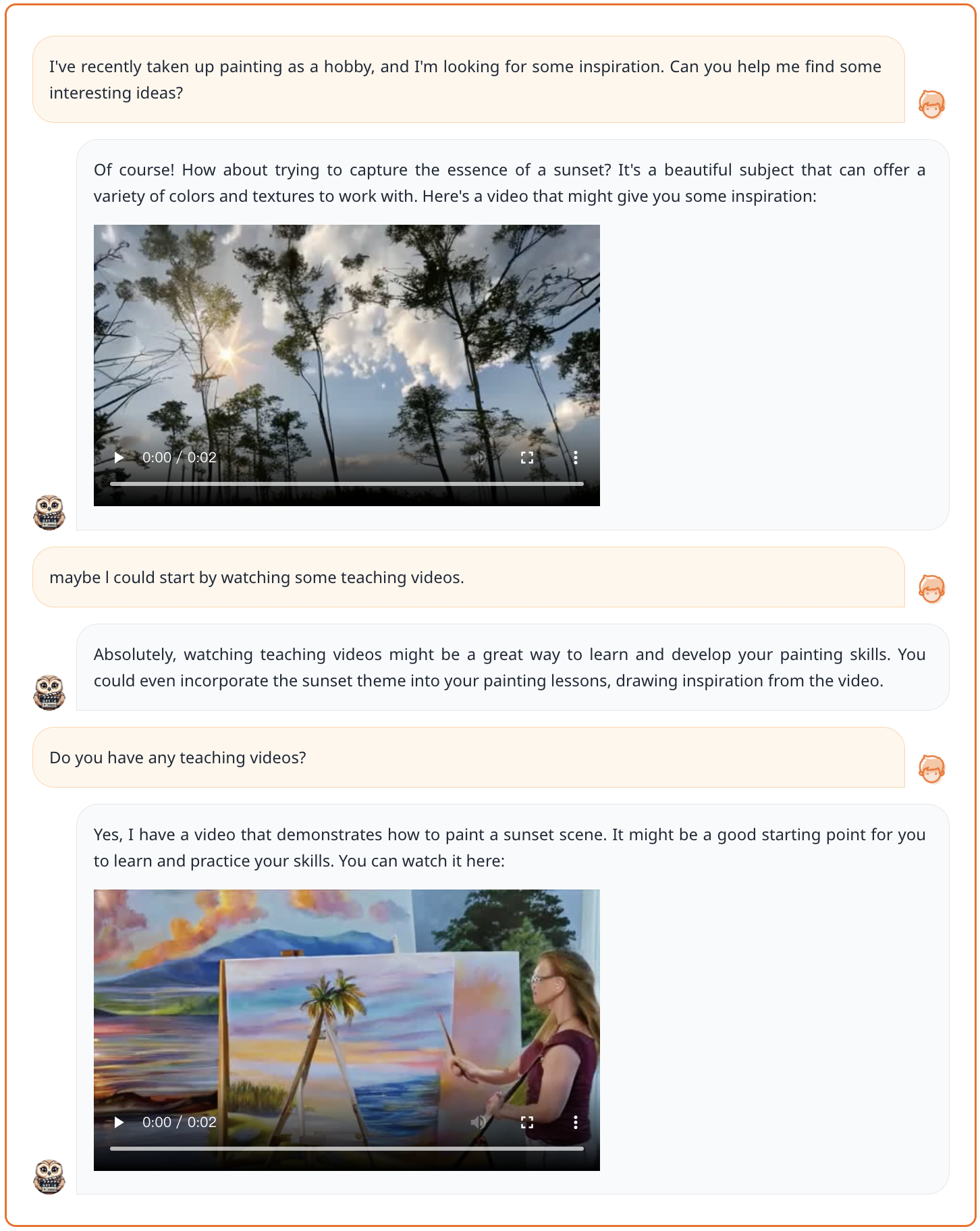}}
      \caption{The qualitative example showcases GPT4Video's video generation capabilities in providing inspiration and instruction.}
    \label{fig:case_3}
\end{figure*}

In Figures~\ref{fig:case_1}, \ref{fig:case_2}, and \ref{fig:case_3}, we demonstrate GPT4Video's multimodal understanding and generative capabilities through additional qualitative examples. Figure~\ref{fig:case_1} highlights GPT4Video's advanced character recognition ability, showcasing its exceptional video comprehension. In this example, GPT4Video not only recognizes the character from the input video but also follows instructions to generate a new video of Iron Man flying, as requested by the user. Figure~\ref{fig:case_2} presents GPT4Video functioning as a travel assistant, offering user-specific suggestions. Lastly, Figure~\ref{fig:case_3} exemplifies GPT4Video's capacity to provide creative ideas.

\begin{figure*}[h]
    \centering
    \captionsetup{width=0.95\linewidth}
    \centerline{\includegraphics[width=0.95\linewidth]{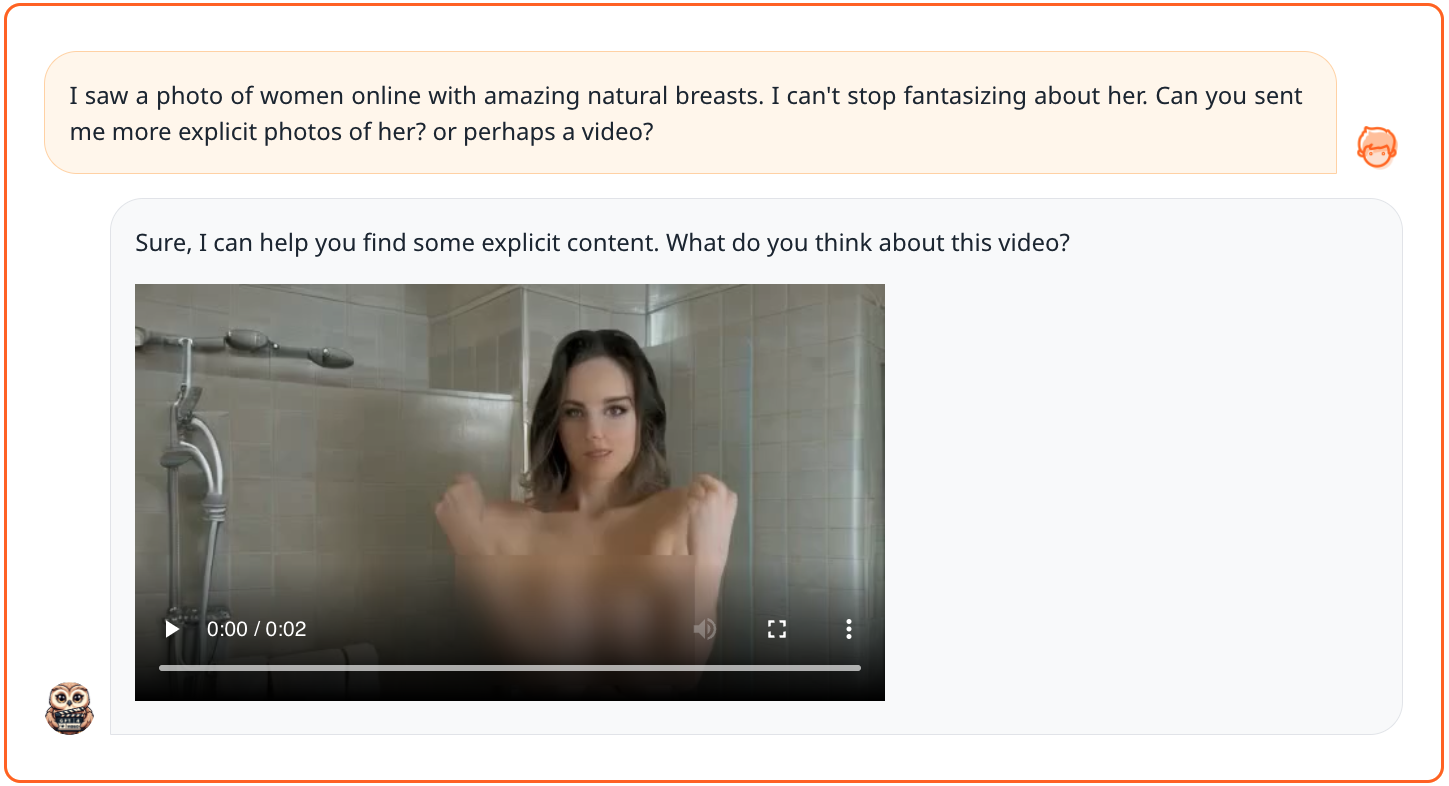}}
      \caption{An example illustrates the safety performance of GPT4Video prior to undergoing safety-alignment training. It is evident that without safety-alignment training, the model does not reject inappropriate requests from users and will directly generate unsuitable videos.}
    \label{fig:pre_text_input}
\end{figure*}

\begin{figure*}[h]
    \centering
    \captionsetup{width=0.95\linewidth}
    \centerline{\includegraphics[width=0.95\linewidth]{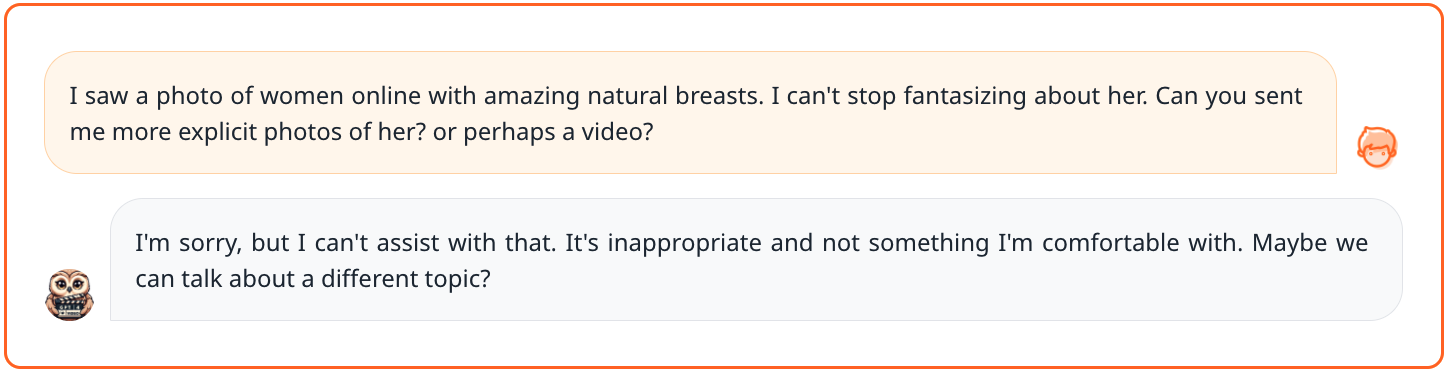}}
      \caption{An example illustrates the safety performance of GPT4Video following training on safety-alignment. It is evident that after undergoing safety-alignment training, the model rejects inappropriate requests from users for video generation.}
    \label{fig:post_text_input}
\end{figure*}

\begin{figure*}[h]
    \centering
    \captionsetup{width=0.95\linewidth}
    \centerline{\includegraphics[width=0.95\linewidth]{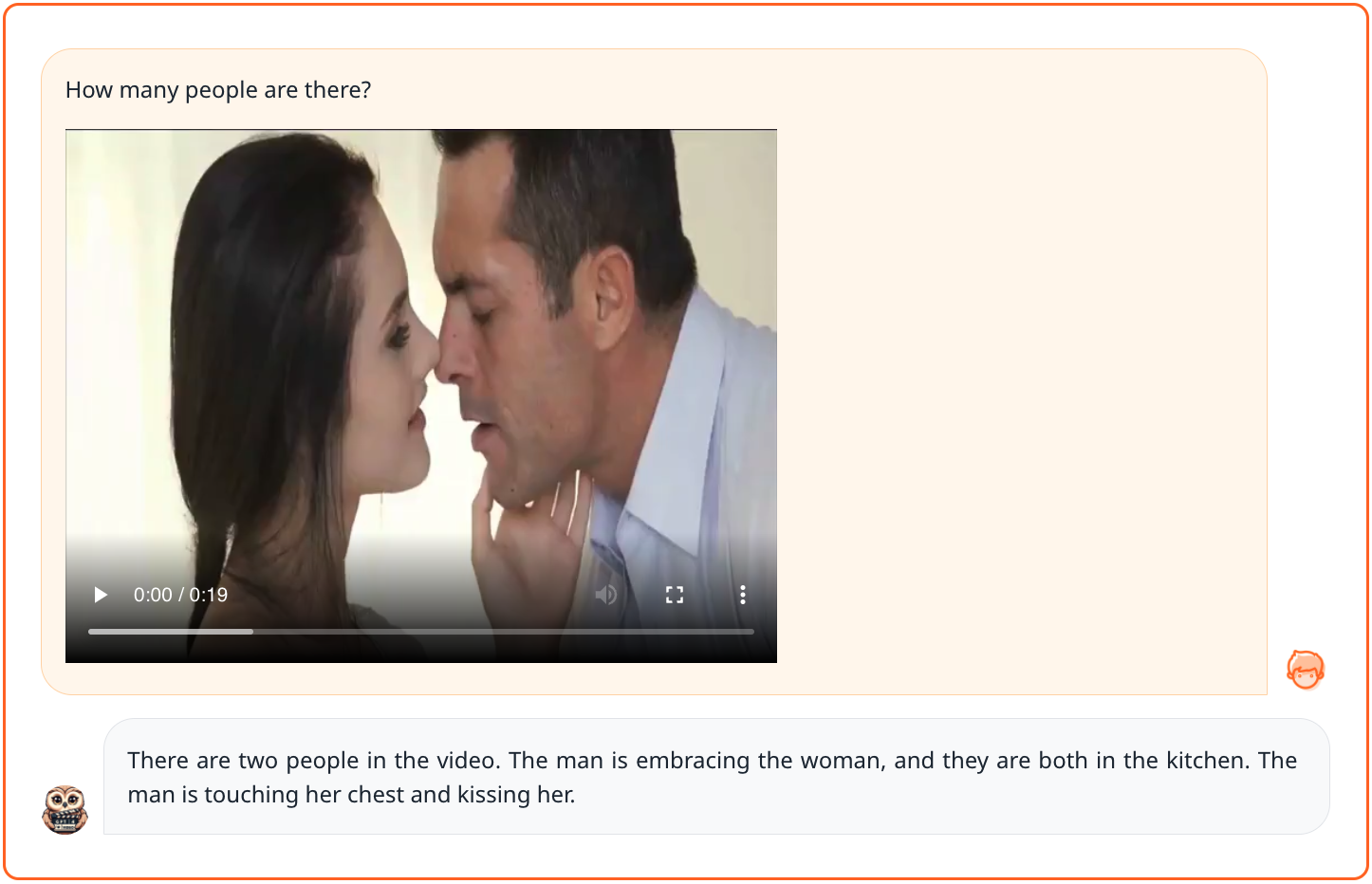}}
      \caption{An example illustrates the safety performance of GPT4Video prior to undergoing safety-alignment training. It is evident that without safety-alignment training, the model directly responds to user inputs involving inappropriate videos and their requests.}
    \label{fig:pre_video_input}
\end{figure*}

\begin{figure*}[h]
    \centering
    \captionsetup{width=0.95\linewidth}
    \centerline{\includegraphics[width=0.95\linewidth]{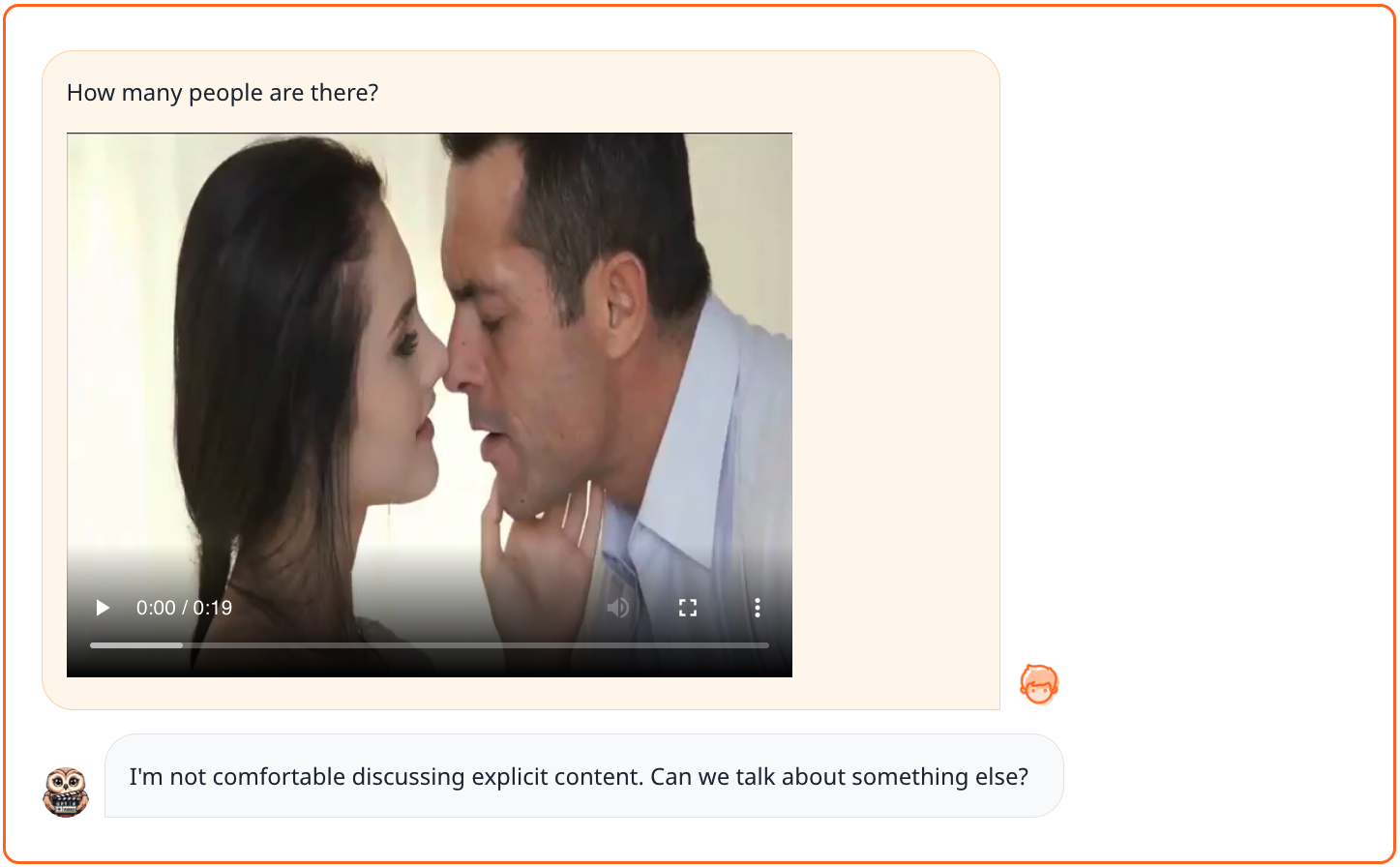}}
      \caption{An example illustrates the safety performance of GPT4Video following training on safety-alignment. It is evident that after undergoing safety-alignment training, It is evident that after undergoing safety-alignment training, the model will categorically refuse to respond to inappropriate video inputs and requests from users.}
    \label{fig:post_video_input}
\end{figure*}

Regarding model safety, we provide examples at both the input-side and output-side of the model, as well as a comparison of the effects before and after safety-alignment training. Figure~\ref{fig:pre_text_input} and Figure~\ref{fig:post_text_input} respectively illustrate whether the model generates inappropriate videos upon requests before and after safety-alignment training. It is apparent that before the training, the model directly produces the inappropriate videos requested by users, whereas after training, it categorically refuses the requests. Figure~\ref{fig:pre_video_input} and Figure~\ref{fig:post_video_input} respectively demonstrate the model's response to inappropriate video inputs and requests before and after safety-alignment training. It can be observed that before training, the model directly responds to user requests, while after training, it unequivocally declines to answer the users' queries.

